\documentclass[10pt,twocolumn,letterpaper]{article}

\usepackage{cvpr}              

\usepackage{times}
\usepackage{epsfig}
\usepackage{graphicx}
\usepackage{amsmath}
\usepackage{amssymb}

\usepackage{graphicx}
\usepackage{amsmath}
\usepackage{amssymb}
\usepackage{booktabs}
\usepackage{textcomp}
\usepackage{comment}
\usepackage{adjustbox}
\usepackage{float}

\usepackage{svg}
\usepackage{tabularx}
\usepackage{pifont}
\newcommand{\cmark}{\ding{51}}%
\usepackage{multirow}

\def\datasetname{HouseCat6D\xspace}
\renewcommand\footnotemark{}

\definecolor{cvprblue}{rgb}{0.21,0.49,0.74}
\usepackage[pagebackref,breaklinks,colorlinks,citecolor=cvprblue]{hyperref}

\title{HouseCat6D - A Large-Scale Multi-Modal Category Level 6D Object Perception Dataset with Household Objects in Realistic Scenarios
}

\author{
HyunJun Jung$^{1 \ast}$, 
Guangyao Zhai$^{1 \ast}$,
Shun-Cheng Wu$^{1 \ast}$, 
Patrick Ruhkamp$^{1 \ast}$, 
Hannah Schieber$^{1,2 \ast}$,\\
Giulia Rizzoli$^{3}$,
Pengyuan Wang$^{1}$,
Hongcheng Zhao$^{1}$,
Lorenzo Garattoni$^{4}$,
Sven Meier$^{4}$,\\
Daniel Roth$^{1}$, 
Nassir Navab$^{1}$,
Benjamin Busam$^{1,5}$\\[0.5em]
$^1$ Technical University of Munich\qquad
$^2$ FAU Erlangen-Nürnberg \qquad
$^3$ University of Padova\\
$^4$ Toyota Motor Europe \qquad
$^5$ 3dwe.ai\\[0.5em]
{\tt\small \{hyunjun.jung,guangyao.zhai,b.busam\}@tum.de}
\vspace{-1cm}
\thanks{*Authors with equal contributions.}
}

\begin{document}

\makeatletter
\g@addto@macro\@maketitle{
  \begin{figure}[H]
  \setlength{\linewidth}{\textwidth}
  \setlength{\hsize}{\textwidth}
  \centering
    \includegraphics[width=\linewidth]{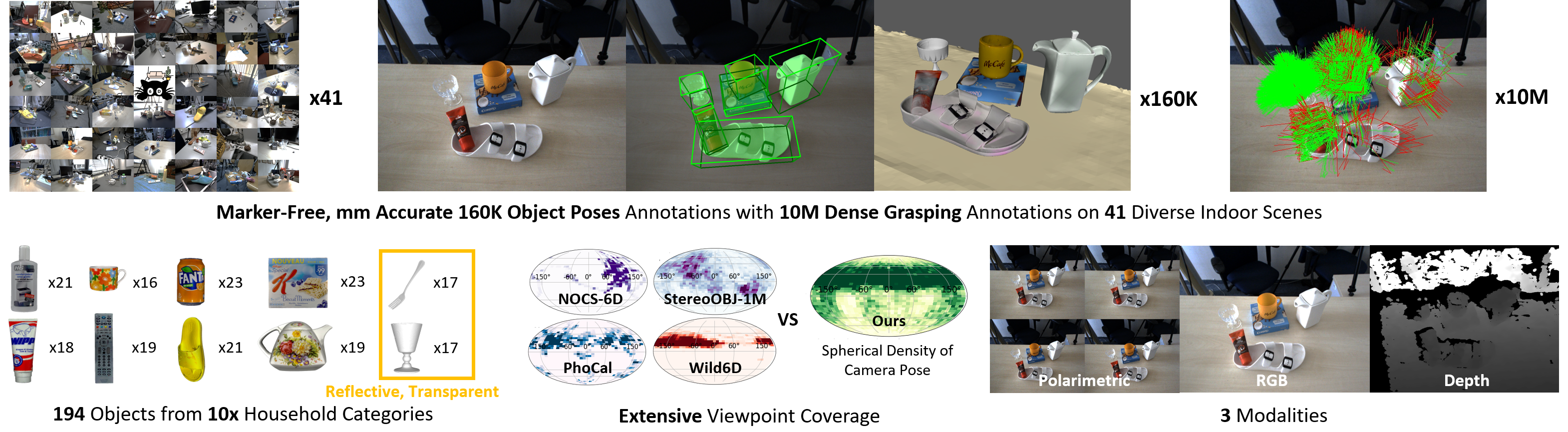}
    \vspace{-5mm}
	\caption{\datasetname~is a multi-modal category level 6D object pose and grasping dataset with highly diverse household object categories of different photometric complexity and a high number of varying scenes covering large viewpoint distributions. 
    It comprises room-scale high-quality camera trajectories and object poses without markers in realistic scenarios including occlusions as well as dense grasping pose annotation.
    Data includes synchronized RGB, depth from active stereo, and polarimetric RGB+P images in scenes comprising objects without texture, strong reflections, or translucency. The project website is \url{https://sites.google.com/view/housecat6d}.
    }
    \label{fig:teaser}
	\end{figure}
}
\makeatother

\maketitle

\begin{abstract}
Estimating 6D object poses is a major challenge in 3D computer vision. Building on successful instance-level approaches, research is shifting towards category-level pose estimation for practical applications. Current category-level datasets, however, fall short in annotation quality and pose variety. Addressing this, we introduce \textbf{\datasetname}, a new category-level 6D pose dataset. It features \textbf{1}) multi-modality with Polarimetric RGB and Depth (RGBD+P), \textbf{2}) encompasses 194 diverse objects across 10 household categories, including two photometrically challenging ones, and \textbf{3}) provides high-quality pose annotations with an error range of only 1.35 mm to 1.74 mm. The dataset also includes \textbf{4}) 41 large-scale scenes with comprehensive viewpoint and occlusion coverage, \textbf{5}) a checkerboard-free environment, and \textbf{6}) dense 6D parallel-jaw robotic grasp annotations. Additionally, we present benchmark results for leading category-level pose estimation networks.
\end{abstract}

\begin{table*}[!t]
\caption{\textbf{Dataset Overview.} \datasetname represents a large-scale and highly accurate category-level 6D pose dataset that combines the advantages of various established datasets (e.g. extensive pose coverage, highly accurate GT, occlusions cases, and grasping annotation).
}%
\resizebox{\textwidth}{!}{

\centering

\begin{tabular}{l | c c c | c c c | c c c c c| c c c | r r r }
\toprule
  \small Dataset &
  \rotatebox[origin=c]{90}{\small RGB}          &
  \rotatebox[origin=c]{90}{\small Depth}        &
  \rotatebox[origin=c]{90}{\small Polarisation} &
  \rotatebox[origin=c]{90}{\small Real}         &
  \rotatebox[origin=c]{90}{\small Multi-View}   &
  \rotatebox[origin=c]{90}{\small mm-accurate GT}   &
  \rotatebox[origin=c]{90}{\small Occlusion}    &
  \rotatebox[origin=c]{90}{\small Symmetry}     &
  \rotatebox[origin=c]{90}{\small Transparent}  &
  \rotatebox[origin=c]{90}{\small Reflective}   &
  \rotatebox[origin=c]{90}{\small Grasping} &
    \rotatebox[origin=c]{90}{\small Pose Density}   &
    \rotatebox[origin=c]{90}{\small Pose Variation}   &
    \rotatebox[origin=c]{90}{\small Workspace}   &
  \rotatebox[origin=c]{90}{\small Categories}   &
  \rotatebox[origin=c]{90}{\small Objects}      &
  \rotatebox[origin=c]{90}{\small Scenes}     \\ 
\midrule
FAT~\cite{tremblay2018falling}                            & \cmark & \cmark &        &          & \cmark &   \cmark     & \cmark & \cmark &        &          & &&&& --     & $21$   & $>1$k \\
BlenderProc~\cite{denninger2020blenderproc}               & \cmark & \cmark &        &          & \cmark & \cmark       & \cmark & \cmark &        &          &&&&& --     & --     & $>1$k \\
LabelFusion~\cite{marion2018label}                        & \cmark & \cmark &        & \cmark   &        &        & \cmark &        &        &          & &&&& --     & $12$   & $138$ \\
Linemod~\cite{hinterstoisser2011multimodal,brachmann2014} & \cmark & \cmark &        & \cmark   &        &        & \cmark & \cmark &        &          & &&&& --     & $15$   & $15$ \\
Toyota Light~\cite{hodan2018bop}                          & \cmark & \cmark &        & \cmark   &        &        &        & \cmark &        &          & &&&& --     & $21$   & $21$ \\
YCB~\cite{calli2015benchmarking,xiang2018posecnn}         & \cmark & \cmark &        & \cmark   &        &        & \cmark & \cmark &        &          & &&&& --     & $21$   & $92$ \\
T-LESS~\cite{hodan2017t}                                  & \cmark & \cmark &        & \cmark   &        &        & \cmark & \cmark &        &          &&&&& --     & $30$   & $20$ \\
HomebrewedDB~\cite{kaskman2019homebreweddb}               & \cmark & \cmark &        & \cmark   &        &        & \cmark & \cmark &        &          &&&&& --     & $33$   & $13$ \\
ITODD~\cite{Drost_2017_ICCV}                              &        & \cmark &        & \cmark   & \cmark &\cmark& \cmark & \cmark &        & (\cmark) & &&&& --     & $28$   & $800$ \\
GraspNet-1Billion~\cite{fang2020graspnet}                 & \cmark & \cmark &        & \cmark   & \cmark    &        & \cmark & \cmark &        &        & \cmark & +++&+++&+& --     & $88$   & $190$ \\
HOPE~\cite{tyree20226} & \cmark & \cmark &  & \cmark   & \cmark &  & \cmark & (\cmark) &  & & &++++&++++&+++& --    & $28$   & $50$\\
StereoOBJ-1M~\cite{liu2021stereobj}                       & \cmark &        &        & \cmark   & \cmark & \cmark  & \cmark & \cmark & \cmark & \cmark  & &++++&++++&+++& --     & $18$   & $183$  \\
\midrule
kPAM~\cite{manuelli2019kpam}                              & \cmark & \cmark &        & \cmark   &        &        & \cmark & \cmark &        &         &&&&& $2$    & $91$   & $362$ \\
TOD~\cite{liu2020keypose}                                 & \cmark & \cmark &        & \cmark   & \cmark & \cmark &        & \cmark & \cmark &         &&&&& $3$    & $20$   & $10$ \\
Wild6D~\cite{Fu2022Wild6D}&\cmark&\cmark&&\cmark&\cmark&& &\cmark&(\cmark)&&&++&++&++++&$5$&$162$&$486$\\
NOCS~\cite{wang2019normalized} & \cmark & \cmark & & & \cmark &&&&&&&+&+&++&$6$&$42$&$18$\\
PhoCaL~\cite{wang2022phocal}                                             & \cmark & \cmark & \cmark & \cmark   & \cmark & \cmark & \cmark & \cmark & \cmark & \cmark &  &+++&+++&+& $8$    & $60$   & $24$\\
\textbf{\datasetname} (Ours) & \cmark & \cmark & \cmark & \cmark   & \cmark & \cmark & \cmark & \cmark & \cmark & \cmark  & \cmark &\textbf{+++++}&\textbf{+++++}&\textbf{+++++}& \textbf{10}    & \textbf{194}   & $41$
\end{tabular}
}

\label{tab:dataset_comparison_gy}%

\end{table*}
\section{Introduction}\label{sec:intro}%
6D pose estimation is one of the cornerstones in many computer vision tasks, especially for interactions like robotic manipulation~\cite{wang2021demograsp,zhai20222,zhai2022monograspnet,zhai2023sg} or augmented reality~\cite{esposito2016multimodal}. 
Many methods have been proposed to solve this task from various perspectives and achieve outstanding results on public benchmarks~\cite{hinterstoisser2011multimodal,kaskman2019homebreweddb,calli2015benchmarking,xiang2018posecnn}. Most of the methods focus on instance-level where each network is trained and tested on a single object instance~\cite{manhardt2019explaining,wang2021gdr}. However, generalization and applicability are limited, as the object mesh is required, and an individual network needs to be trained for each instance. 
Recent methods focus on category-level pose estimation~\cite{wang2019normalized, manhardt2020cps, manhardt2020cps++, lin2021dualposenet, chen2021fs} by training on multiple objects within one category. They can later generalize to unseen objects from the same category. 
However, a significant limitation blocking further progress is the lack of datasets for training and evaluation that fulfill all criteria like large-scale, accurate, and realistic. 
Existing category-level datasets only comply partly, e.g., high quantity and low quality~\cite{wang2019normalized}, or high quality but insufficient quantity~\cite{wang2022phocal}. 

To this end, we propose a new category-level dataset \datasetname. It consists of high-quality ground-truth annotations on diverse objects acquired by multiple sensor modalities with extensive viewpoint coverage. Our dataset includes 194 objects from 10 different categories, including photometrically challenging classes such as glass and cutlery (\cref{fig:teaser}), occlusion cases, and 3 sensor modalities, i.e., RGB, depth, and polarimetric images, with a total of 23.5k frames and approx. 160k annotated object poses. %
We additionally provide 10M grasp pose annotations to a subset of the dataset, endowing it with the capacity to serve robotic manipulation tasks, e.g., category-level robotic grasping~\cite{wen2022transgrasp}.
Our dataset recording relies on an accurate external infrared tracking system and additional subsequent post-processing through sparse bundle adjustment to avoid errors induced by timestamp offsets and motion blur of the freely moving camera rig~\cite{colmap_1,colmap_2}. %
Specifically, we conduct three calibrations, \ie pivot calibration, timestamp calibration, and hand-eye-calibration. 
For the timestamp calibration, we adopt existing methods~\cite{ar_sync1,ar_sync2} adjusted to our setup with an ICP-based refinement. For the hand-eye-calibration, we improve the calibration from recent work~\cite{wang2022phocal} by aggregating multiple measurements of a ChArUcO~\cite{an2018charuco} calibration board (\cref{sec:camera_trajectory_annotation}). 
Compared to the recent PhoCaL dataset~\cite{wang2022phocal} that relies on a robotic end-effector to estimate poses and thus has limited viewpoint coverage and backgrounds, our method provides accurate object pose annotation and wide viewpoint coverage while providing pose annotations of similar quality. 
We use active stereo as depth maps, which is more reliable on different surface materials~\cite{wang2022phocal, hammer}. In addition to the typical RGB and depth, we provide polarimetric images with four different filter angles. Recent investigations have shown that this modality is especially suitable for tasks such as depth and surface normal estimation~\cite{verdie2022cromo,kalra2020deep, polarimetric_in_wild}, and 6D pose estimation~\cite{polarimetric_pose}, especially for photometrically challenging objects or surfaces.
In summary, our main contributions are: 
\begin{enumerate}
\item We propose \datasetname~a \textbf{large-scale multi-modal category-level object pose dataset} with RGBD + RGBP data, comprising 194 high-quality 3D models of household objects including transparent and reflective objects in 41 scenes with broad viewpoint coverage, challenging occlusions and no markers. 
\item We develop a novel \textbf{pipeline for annotation, recording, and post-processing} to achieve comparable accuracy to robotic GT, but with a mobile handheld multi-camera rig. We detail all acquisition and calibration steps and make the \textbf{high-quality 6D object pose annotations together with 6D grasp labels} accessible to the community. 
\item We provide and discuss the \textbf{benchmark evaluation results} on \datasetname for SOTA category-level baselines to show challenges and foster novel research in the field. 
\end{enumerate}

\begin{figure*}[t!]
 \centering
    \includegraphics[width=1.0\linewidth]{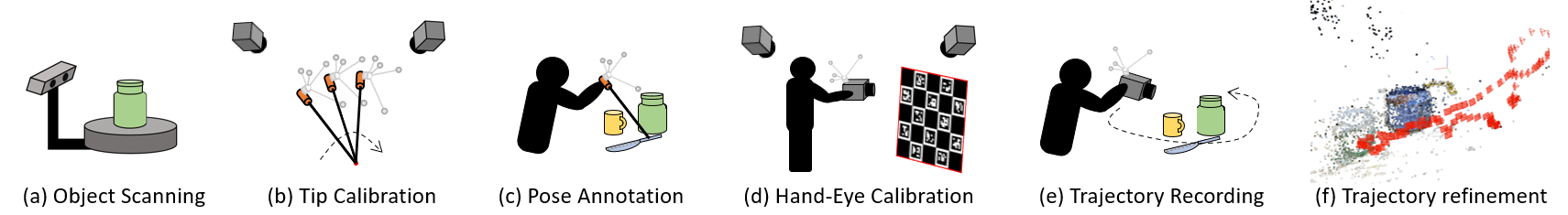}
    \vspace{-2mm}
    \caption{\textbf{Dataset Acquisition Pipeline.} (a): Pre-scanning 3D models. (b): Pivot calibration to calibrate measurement tip from the tracking body. (c): Pose annotation of objects using measurement tip. (d): Hand-Eye-Calibration to calibrate camera center of tracking body. (e): Camera trajectory recording (f): Post-processing step to reduce synchronization-induced trajectory error.}
    \label{fig:pipeline}
\end{figure*}
\section{Related Work}
\label{sec:related}
The recent state-of-the-art methods are mostly data-driven approaches. A common need for these methods is a dataset for training and evaluation. In this section, we give an overview of existing datasets and provide a summary of mentioned datasets in \cref{tab:dataset_comparison_gy}.

\subsection{Instance-level 6D Object Pose Dataset}
\label{subsec:related-instance}
Early-stage datasets provide nontemporal consistent images. LineMOD~\cite{hinterstoisser2011multimodal} and LM-Occlusion~\cite{brachmann2014} are arguably the most used datasets. They use an RGBD camera to annotate the pose of the objects. The camera pose is estimated with checkerboards, which constantly appear in all images. Although these two datasets were heavily used, the quality of object meshes and annotations varies~\cite{busam2020like}. Other datasets were proposed to overcome these issues. Such as HomebrewedDB~\cite{kaskman2019homebreweddb} and others ~\cite{doumanoglou2016recovering, tejani2014latent}. However, those datasets still rely on checkerboard-based camera localization, or human-powered annotation~\cite{2016dataset}, or a rotating table~\cite{Drost_2017_ICCV} to provide tolerable annotations.
Consecutive datasets focus on providing sequential images with camera and object pose annotations. This allowed to investigate pose tracking approaches with temporal constraints~\cite{busam2020like, laval_error, Li_2019_iterative}. One very popular such benchmark is YCB~\cite{xiang2018posecnn}. 
The annotation is achieved by leveraging an RGBD camera and Structure from Motion (SfM)~\cite{marion2018label}. Although this makes large-scale annotation possible, the annotation quality is bound to the quality of the depth camera used~\cite{wang2022phocal,hammer}. In comparison, the Laval 6DOF dataset~\cite{laval_error} marker-based tracking results in high-quality annotations and checkerboard-free images. However, marker-induced depth artifacts need a depth map post-correction. On the other hand, StereoOBJ-1M~\cite{liu2021stereobj} uses SfM with checkerboards in a more precise way to ensure quality and quantity. However, this also introduces checkerboards in every image.
GraspNet-1Billion~\cite{fang2020graspnet} provides parallel-jaw grasping labels besides object pose annotations, making it more feasible for downstream robotic bin-picking. However, the dataset has limited viewpoint changes and only simple backgrounds.
In contrast, our dataset captures multiple household scenarios with adequate viewpoint coverage.\par%
\begin{figure*}[!t]
 \centering
    \includegraphics[width=1.0\linewidth]{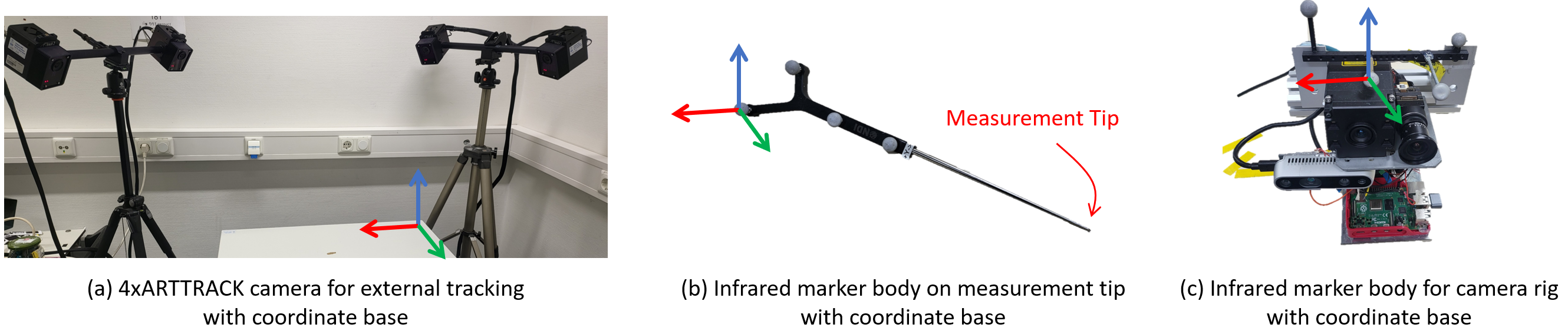}
    \caption{\textbf{Tracking System.} ARTTRACK2 tracking system and sets of infrared marker bodies we used for our setup. Once at least four infrared spheres are detected from at least two cameras, the tracking system provides the pose of the marker body as transformation from tracker system base to marker body base.}
    \label{fig:setup_example}
\end{figure*}
\subsection{Category-level 6D Object Pose Dataset}
\label{subsec:related-category}
Category-level pose estimation has been proposed to address generalizability in 6D pose estimation over multiple objects of the same category. The task is to generalize pose estimation per class and not for individual instances, which is challenging due to high intra-class variance. Many recent methods have been proposed~\cite{Chen_2021_CVPR, di2022gpv,tian2020shape,manhardt2020cps,manhardt2020cps++,centersnap,chen2020learning,lin2021dualposenet} to solve this problem due to its realistic setup. Only a few datasets exist, which we will briefly review here.\par%
The NOCS dataset~\cite{wang2019normalized} is the first category-level 6D pose dataset. It contains six categories and two sub-datasets, namely CAMERA25 and REAL275. In REAL275, the poses are aligned using checkerboards. For CAMERA25, ShapeNetCore~\cite{chang2015shapenet} objects are placed in table-top scenarios. 
A dataset focusing more on the robotic field is kPam, which uses keypoints. Manuelli \etal~\cite{manuelli2019kpam} capture kPam using a similar approach as Marion et al.~\cite{marion2018label}. They perform 3D reconstruction before manually labeling the keypoints on the 3D reconstruction. The dataset results in 117 training sequences and 245 testing sequences. 
While NOCS and kPAM contain solid objects, TOD~\cite{liu2020keypose} and PhoCaL~\cite{wang2022phocal} specifically focus on either translucent or transparent and reflective objects. TOD~\cite{liu2020keypose} is captured with a robotic arm and annotated keypoints and focuses on stereo images. Wang \etal~\cite{wang2022phocal} introduce a category-level dataset including polarimetric images besides RGBD only. For annotation, a robotic arm is used to tip individual objects with a calibrated pointer. Annotations are refined via ICP. 
Instead of using a robotic arm, Wild6D~\cite{Fu2022Wild6D} is annotated via tracking. Every 50th keyframe is annotated and then registered via TEASER++~\cite{yang2020teaser} and colored ICP~\cite{park2017colored}. The training dataset is label-free, and only the test dataset contains annotations. For recording, multiple iPhones are used to capture RGB images, depth, and point cloud. Fu \etal~\cite{Fu2022Wild6D} introduce a Wild6D, an unlabeled RGBD video dataset with diverse scenes. They also investigate the use of additional synthetic labels and annotate a fraction of the real videos for evaluation.
Other class-based datasets exist. Objectron~\cite{ahmadyan2021objectron} focuses on scale and provides over 14k scenes. However, it only provides annotated 3D bounding boxes and does not give detailed shape information for the objects.

\section{Dataset}\label{sec:dataset}

\begin{figure}[!h]
 \centering
    \includegraphics[width=1.0\linewidth]{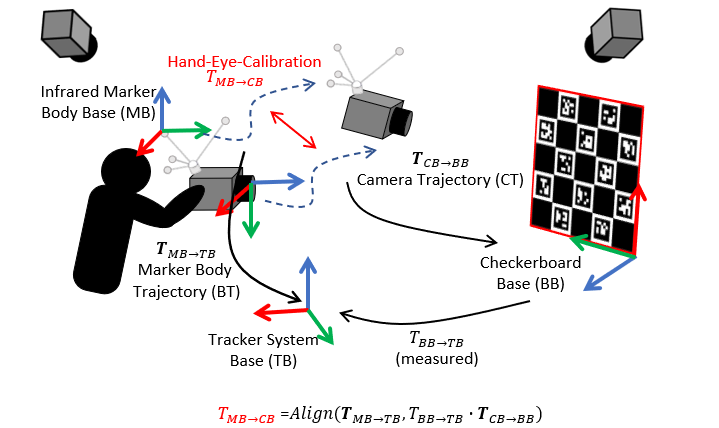}
    \caption{\textbf{Hand-Eye-Calibration.} Instead of single image based using closed form solution with measured checkerboard pose like in~\cite{wang2022phocal}, our newly proposed approach takes more image captures into account. This makes it more robust against checkerboard detection errors in a wider range of camera poses.}
    \label{fig:handeye_calibration}
\end{figure}
Our dataset aims to provide large scale with extensive view coverage and high-quality pose annotations without a checkerboard. It is composed of 34 training scenes (20k frames), five test scenes (3k frames), and two validation scenes (1.4k frames). The scenes comprise objects from 10 household categories, including photometrically challenging objects like glass and cutlery, with occlusions. With a total of 194 objects, each category contains 19 objects on average. Our dataset also features multiple modalities, namely RGB images, polarimetric images, and depth maps. This section details our dataset. The acquisition setup is described in \cref{fig:pipeline}.\par%
\subsection{Objects \& Hardware}\label{sec:hardware}
Here, we briefly describe the hardware setup we use for the dataset acquisition. More detailed information, such as product names and their specs, is provided in the supplementary material.
For our dataset, we choose 10 household categories to represent typical household scenarios: bottle, box, can, cup, cutlery, glass, remote, shoe, teapot, and tube. All objects are scanned with a structured light stereo-based 3D scanner to ensure the quality of the reconstructed meshes. For the photometrically challenging categories, we use self-vanishing 3D scanning spray to enable scanning. For tracking the annotation tool and camera rig, we utilize an external tracker system composed of 4 infrared cameras to ensure tracking quality without using a checkerboard. \cref{fig:setup_example} shows our camera setup and used tracking bodies for the annotation pipeline. We evaluate the accuracy of the tracking with translation and rotational error~\cite{laval_error} using a robotic setup (details in supplementary material). The average error is 0.67 \text{mm} / 0.12\textdegree~in the static case and 0.92 \text{mm} / 0.16\textdegree~in the dynamic tracking scenario. Our dataset comprises two main modalities: Polarimetric RGB image and active stereo depth. We use a dedicated sensor for each of the modalities. For polarimetric images, we use a polarimetric camera, which produces four polarized RGB images for every shot. To measure depth, we decided on an active stereo depth sensor over Time-of-Flight sensors as active stereo depth provides, in general, more robust depth on photometrically challenging material~\cite{hammer}. To synchronize the two cameras, an external hardware trigger is used to trigger both cameras simultaneously.\par%
\begin{figure*}[!ht]
 \centering
    \includegraphics[width=1.0\linewidth]{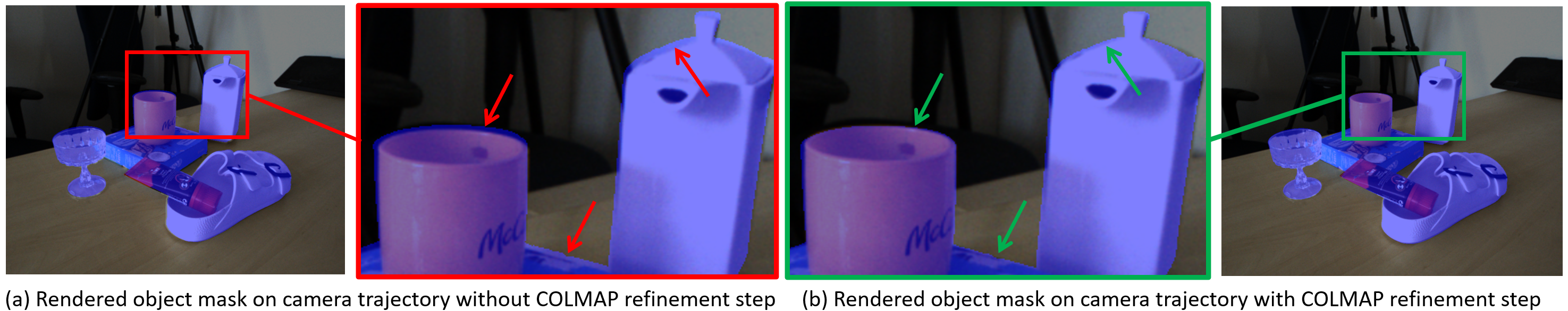}
    \vspace{-4mm}
    \caption{\textbf{Post-Processing via Bundle Adjustment.} Example of COLMAP~\cite{colmap_1,colmap_2} refinement on selected frame with large camera displacement. Even though our timestamp synchronization step reduces the effect of motion induced pose offset, subtle errors still remain ((a), marked red). In comparison, the post processing step significantly reduces the given offset error ((b), marked green).}
    
    \label{fig:colmap_refinement}
\end{figure*}
\subsection{Object Pose Annotation}\label{sec:pose_annotation}
Annotating the 6D pose of the object is, without a doubt, the most crucial part of a 6D pose dataset. In our dataset, we adopt the highly accurate object pose annotation pipeline from~\cite{wang2022phocal} but replace the robotic end-effector pose with an IR tracking body. This ensures reliable tracking quality while covering a more extensive working volume. The annotation step follows tool tip calibration, 3D points measurement of the objects, and point correspondence with ICP-based refinement. In this subsection, we describe the details of each step.\par%
\paragraph{Tip Calibration.}
The poses of the object meshes are annotated by measuring the 3D point using the tooltip. Thus, calibrating the location of the tip from the tracking body is essential to ensure the accuracy of the annotation. We use an NDI Active 4-Marker Planar Rigid Body (Northern Digital, Ontario Canada) as the measurement tip (\cref{fig:setup_example} (b)). The tip is calibrated by fixing the tip while pivoting the tracking body and finding the optimal location of the point to minimize the variance of the fixed point (pivot calibration). The most common way to evaluate the quality of the pivot calibration is by measuring the variance of the fixed pivot point. We carefully calibrated with 18 points, with the final variance of the tip location of $\varepsilon = 0.040~\text{mm}$.\par%
\paragraph{Pose Annotation.}\label{sec:object_pose_annotation}%
After the tip is calibrated from the tracking body, it can measure accurate 3D points in space in the world coordinates of the tracker system. We measure points for the initial point correspondence and ICP refinement as in~\cite{wang2022phocal} while covering around three times more point measurements with various surfaces of the object, thanks to the enlarged working space without the constraint given by using a robot arm~\cite{wang2022phocal}. We evaluate the quality of the pose annotation step by simulating the pose annotation pipeline on randomly selected three objects with the addition of pivot calibration error (\cref{sec:pose_annotation}) and static tracking error (\cref{sec:hardware}), which gives an average RMSE of $0.32$ mm in translation and $0.43^{\circ}$ in rotation.\par%
\subsection{Camera Trajectory Annotation}\label{sec:camera_trajectory_annotation}
Another critical aspect of 6D pose annotation is accurate camera trajectories. The object poses are annotated from the center of the tracker system, not from the the individual camera. Thus, the camera pose from the tracker system base has to be applied to obtain the 6D pose of the object from the camera center. In this section, we describe the detailed steps of camera trajectory annotation precisely.\par%
\paragraph{Hand-Eye-Calibration}\label{sec:hand_eye_calibration}
In our scenario, Hand-Eye-Calibration obtains the transformation between the tracker marker body and the center of the camera image sensor. The most common way to perform Hand-Eye-Calibration~\cite{tsai1989new} is detecting the checkerboard multiple times via camera while tracking the camera body from an external source and optimizing both checkerboard base from the tracker system base \({T}_{BB\rightarrow TB}\) and camera base from the marker body base \({T}_{MB\rightarrow CB}\) (hand-eye-calibration). In comparison, \cite{wang2022phocal} proposes a way to use the measurement tip to measure \({T}_{BB\rightarrow TB}\) and form a close form solution with a single checker board detection to obtain \({T}_{MB\rightarrow CB}\). However, we found that the accuracy of the closed-form solution is often unreliable. To solve this, we propose a new hand-eye calibration, which takes into account multiple image captures (\cref{fig:handeye_calibration}). We capture multiple static images from different locations to form two trajectories - one from the camera and one from the tracking body and extract a fixed offset matrix by applying Horn's alignment method~\cite{horn}, which is the hand-eye-calibration matrix. Once the calibration \({T}_{MB\rightarrow CB}\) is obtained, we align the two trajectories and compare the errors to the calibration accuracy. The RMSE for this calibration is measured as \(0.27\,\text{mm}\) for the translation and $0.42^{\circ}$ for the rotation.\par%
\begin{figure}[t]
 \centering
    \includegraphics[width=1.0\linewidth]{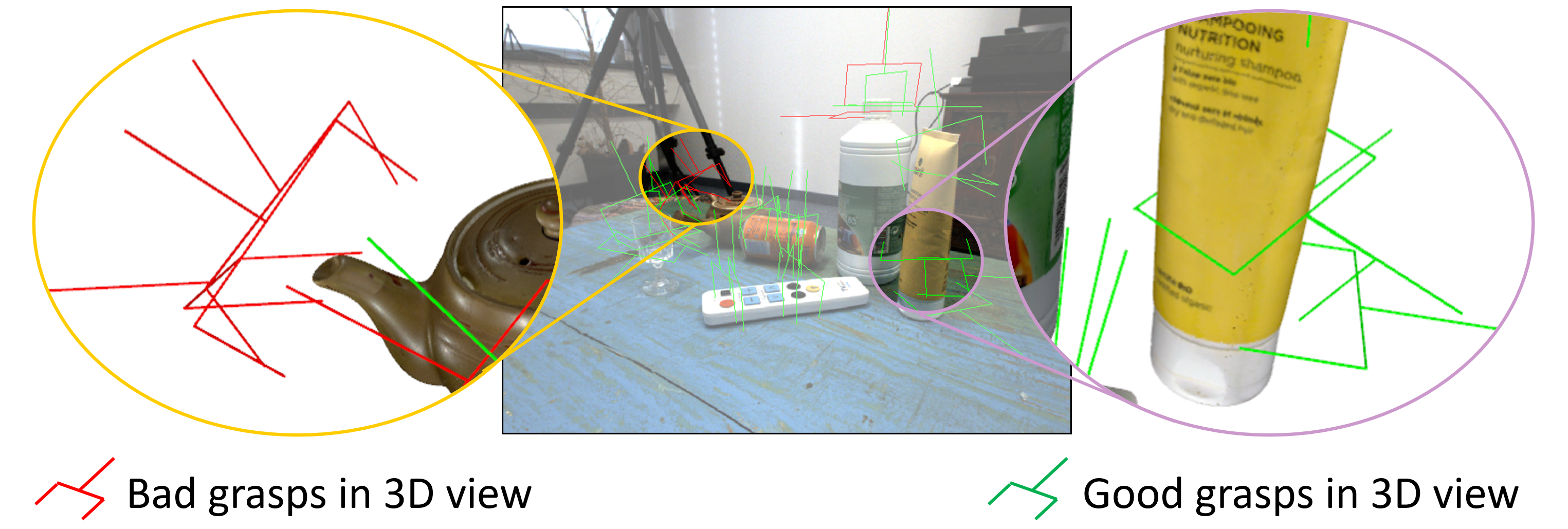}
    \vspace{-4mm}
    \caption{\textbf{Grasp Annotations.} After inspection of grasps, we annotate successful grasps (here coloured in green) and failed grasps (here red). We downsample the amount of annotated grasps for better visualization.}
    \label{fig:grasp}
    \vspace{-5mm}
\end{figure}

\begin{figure*}[!t]
 \centering
    \includegraphics[width=\linewidth]{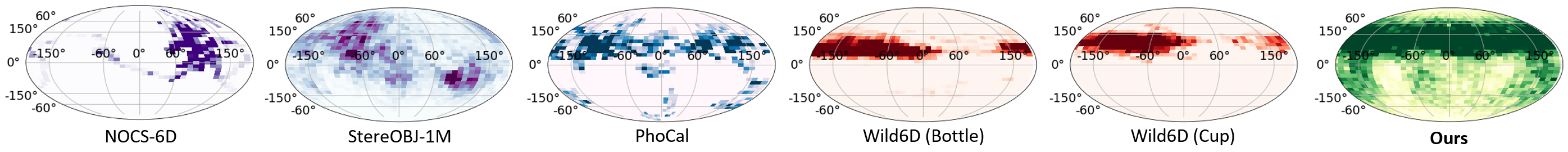}
    \caption{\textbf{Pose Distribution. }
    The pose distribution for category-level datasets NOCS~\cite{wang2019normalized} (Test), StereOBJ-1M~\cite{liu2021stereobj} (Val), PhoCal~\cite{wang2022phocal} (Train), two categories of Wild6D~\cite{Fu2022Wild6D} and Ours is plotted as the Mollweide projection of the spherical histogram, to exemplify the density and pose variation. Ours shows larges diversity of poses around objects, also for the lower hemisphere, and denser overall distribution. 
    }
    \label{fig:histograms_comp}
\end{figure*}%

\begin{figure*}[!htp]
 \centering
    \includegraphics[width=\linewidth]{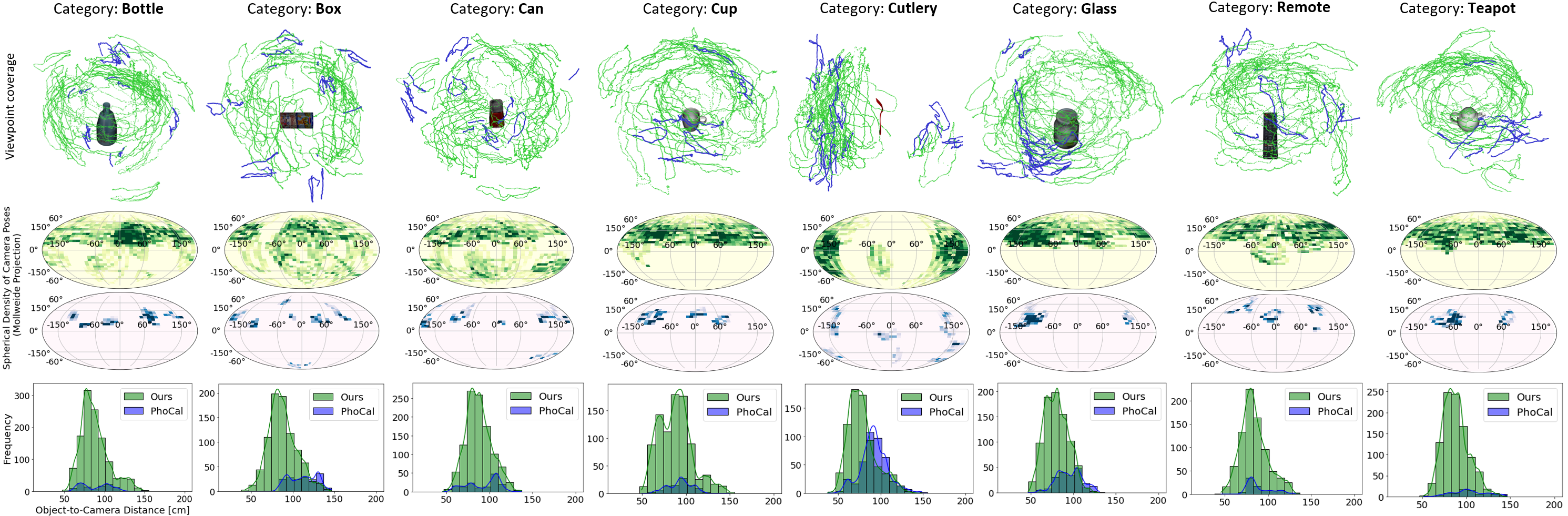}
    \caption{\textbf{Pose Distribution per Category.} We compare the pose distribution of \datasetname (green) against PhoCal~\cite{wang2022phocal} dataset (blue), for the categories included in both of them.
    The trajectory visualization (top) verifies the much larger and better distributed pose coverage of our \datasetname dataset.
    Compared is further the rotational pose coverage as a spherical histogram plotted as Mollweide projection (center) and the object-to-camera distance as a histogram with relative frequency against the distance in cm (bottom).
    }
    \label{fig:histograms}
    \vspace{-5mm}
\end{figure*}%
\paragraph{Camera to Tracker Time Synchronization}
Another important aspect of using the external IR tracking system is the timestamp calibration between the tracking system and the camera image acquisition time. It can cause severe offset on poses depending on the movement of the camera.
A common practice to synchronize the timestamp difference is to measure the trajectory of the camera from two modalities via image and tracking system, along with their timestamp, and maximize the similarity between the trajectories. This brings the best timestamp offset~\cite{ar_sync1, ar_sync2}. In our case, we use ICP-based trajectory alignment to find the best timestamp offset instead of using a similarity measure. We empirically find it is more robust to noise and able to synchronize two trajectories with arbitrary frequency without any interpolation to match the frequency. For the camera timestamp, we use the hardware trigger timestamp.
We evaluate the synchronization by simulating signals with measured noise. One with the tracking system error (\cref{sec:hardware}) and one with a detection-based error (\cref{sec:hand_eye_calibration}). The simulated error is measured as 0.03~sec.\par%

\paragraph{Pose Refinement}
Although time synchronization improves the quality of the camera pose, the motion-induced pose error cannot be obliterated as the time synchronization is imperfect due to noise in the checkerboard detection during calibration~\cite{ar_sync1} as well as the difference in individual camera image acquisition time due to its hardware condition. This effect can be observed when camera motion involves large displacement between consecutive frames (\cref{fig:colmap_refinement}, (a)). To tackle this, we use the RGB input to minimize the reprojection error with multi-view images. We use structure from motion~\cite{colmap_1, colmap_2} with given initial poses and carefully selected fixed frames. The initial poses are used for initial feature matching and structure reconstruction. The fixed frames are excluded in the later bundle adjustment stage. These frames are manually picked upon careful inspection of the frame with the largest IoU between rendered object masks on the RGB image given the pose annotation. We show the improvement in \cref{fig:colmap_refinement}, (b).\par
\begin{table}[!t]
    \centering
    \caption{\textbf{Accuracy Comparison Against Existing Datasets.} While RGBD-based datasets are limited by sensor standard deviation~\cite{liu2021stereobj}, multi-view setups~\cite{liu2020keypose,liu2021stereobj} offer improvements. Our dataset annotation quality, though not as high as robotic acquisitions~\cite{wang2022phocal}, surpasses checkerboard-based datasets~\cite{liu2021stereobj}, and excels in terms of viewpoint coverage and annotation accuracy.
    }
    \resizebox{\columnwidth}{!}{
    \begin{tabular}{l| c c c c c}
    \toprule
    Dataset & RGBD based & TOD~\cite{liu2020keypose} & StereOBJ~\cite{liu2021stereobj} &  PhoCal~\cite{wang2022phocal} & Ours\\
    \hline
    3D Labeling & Depth Map & Multi-View & Multi-View & Robot & IR tracker\\
    Point RMSE [mm] & $\geq 17$ & $3.4$ & $2.3$ & $0.80$ &  $1.35 \leq \epsilon \leq{1.73}$\\
    \bottomrule
    \end{tabular}
    }
    \label{tab:pose_accuracy}
    \vspace{-5mm}
\end{table}

\subsection{Grasp Annotation}\label{sec:grasp_annotation}
To facilitate downstream robotic manipulation tasks, \eg, robotic pick and place, we endow \datasetname with feasible 6D grasping poses for every object under each frame for a subset of collected sequences, following the well-established pipeline introduced in~\cite{fang2020graspnet,eppner2021acronym}. Taking the annotation process for one object as an example: Firstly, we use antipodal sampling with an inspection of Isaac Gym~\cite{makoviychuk2isaac} to distinguish the successful grasps $\mathbf{G}_{obj}^{s}$ from failed ones $\mathbf{G}_{obj}^{uns}$ to generate grasp candidates $\mathbf{G}_{obj}:=\{\mathbf{G}_{obj}^{s}, \mathbf{G}_{obj}^{uns}\}$ for the object mesh. Then we use the obtained object pose under the tracker base \({T}_{obj\rightarrow TB}\) to transform $\mathbf{G}_{obj}$ to the tracker frame where we also reconstruct and amend the background. Finally, we perform collision checking and prune grasping labels in the whole environment and reproject remaining grasps $\mathbf{G}_{TB}$ to each camera frame in the whole sequence according to the transformation from the tracker base to each camera base \({T}_{TB\rightarrow CB}\). An annotated example is shown in \cref{fig:grasp}. More details about the annotation pipeline and parameters can be found in the supplementary material.\par%

\subsection{Pose Annotation Quality Evaluation}\label{sec:annotation_evaluation}
For the evaluation of the annotation quality of object poses, we report the point-wise RMSE between objects and the camera center with and without the consideration of three systematic errors: tracking system error (\cref{sec:hardware}), pose annotation error (\cref{sec:object_pose_annotation}) and hand-eye-calibration error (\cref{sec:hand_eye_calibration}). 
As the accuracy gain from the structure-from-motion cannot be directly quantified, we report the RMSE with upper- and lower bounds. In the upper bound, we report the number with the object annotation error and the static tracking error, assuming no synchronization error. In the lower bound, we include all three mentioned systematic errors, including dynamic tracking error as the tracking system error. We report our annotation quality compared to recent datasets in \cref{tab:pose_accuracy}. Our method achieves a low RMSE of $1.35$ mm to $1.73$ mm.\par%
\begin{table*}[!t]
\caption{\textbf{Quantitative Benchmark Comparisons.} Class-wise evaluation of 3D IoU (at 25\%, 50\%)  for NOCS~\cite{wang2019normalized}, FS-Net~\cite{Chen_2021_CVPR},  VI-Net~\cite{lin2023vi} and GPV-Pose~\cite{di2022gpv} on the test split of \datasetname. Best results on the full set are reported in bold.}
    \centering
    \resizebox{\textwidth}{!}{ 
    \begin{tabular}{l|c|cccccccccc} \toprule
       Approach & $\text{3D}_{25}$ / $\text{3D}_{50}$ & Bottle & Box & Can & Cup & Remote & Teapot & Cutlery & Glass & Tube & Shoe  \\ \midrule
        ~~NOCS~\cite{wang2019normalized} & 50.0 / 21.2 & 41.9 /  5.0 & 43.3 /  6.5 & 81.9 / 62.4 & 68.8 /  2.0 & 81.8 / 59.8 & 24.3 /  0.1 & 14.7 /  6.0 & 95.4 / 49.6 & 21.0 / 4.6 & 26.4 / 16.5 \\ 
        ~~FS-Net~\cite{Chen_2021_CVPR} & 74.9 / 48.0 & 65.3 / 45.0& 31.7 /  1.2 & 98.3 / 73.8 & 96.4 / 68.1 & 65.6 / 46.8 & 69.9 / 59.8 & 71.0 / 51.6 & 99.4 / 32.4 & 79.7 / 46.0 & 71.4 / 55.4\\
        ~~GPV-Pose~\cite{di2022gpv} & 74.9 / 50.7 & 66.8 / 45.6 & 31.4 /  1.1 & 98.6 / 75.2 & \textbf{96.7} / 69.0 & \textbf{65.7 / 46.9} & \textbf{75.4 / 61.6} & 70.9 / 52.0 & \textbf{99.6 / 62.7} & 76.9 / 42.4 & 67.4 / 50.2\\
        ~~VI-Net~\cite{lin2023vi} & \textbf{80.7 / 56.4} & \textbf{90.6 / 79.6} & \textbf{44.8 / 12.7} & \textbf{99.0 / 67.0} & \textbf{96.7 / 72.1} & 54.9 / 17.1 & 52.6 / 47.3 & \textbf{89.2 / 76.4} & 99.1 / 93.7 & \textbf{94.9 / 36.0} & \textbf{85.2 / 62.4}\\
        \bottomrule
    \end{tabular}}
    
    \label{tab:benchmark}
\end{table*}

\begin{table*}[!t]
\caption{\textbf{Sensor Depth Issues and Pose Coverage Influence.} Class-wise evaluation of 3D IoU (at 25\%, 50\%).
NOCS* denotes using ground truth NOCS maps and sensor depth for lifting. VI-Net*~\cite{lin2023vi} denotes a reduced training set. VI-Net is the best baseline. 
}
    \centering
    \resizebox{\textwidth}{!}{ 
    \begin{tabular}{l|c|cccccccccc} \toprule
       Approach & $\text{3D}_{25}$ / $\text{3D}_{50}$ & Bottle & Box & Can & Cup & Remote & Teapot & Cutlery & Glass & Tube & Shoe  \\ \midrule
        ~~VI-Net~\cite{lin2023vi} & 80.7 / 56.4 & 90.6 / 79.6 & 44.8 / 12.7 & 99.0 / 67.0 & 96.7 / 72.1 & 54.9 / 17.1 & 52.6 / 47.3 & 89.2 / 76.4 &  \textbf{99.1 / 93.7} & \textbf{94.9 / 36.0} & 85.2 / 62.4\\
        ~~NOCS*~\cite{wang2019normalized} & \textbf{96.7} / \textbf{93.6} & \textbf{99.8}/ \textbf{98.2}  & \textbf{98.3} / \textbf{95.8}  & \textbf{100.0} / \textbf{99.0} & \textbf{100.0} / \textbf{99.0}  & \textbf{100.0} / \textbf{97.3} & \textbf{99.9} / \textbf{86.1} & \textbf{100.0} / \textbf{99.9} & 80.0 / 68.0 & 89.1/29.1 & \textbf{99.7} / \textbf{83.9} \\  
        ~~NOCS~\cite{wang2019normalized} & 50.0 / 21.2 & 41.9 /  5.0 & 43.3 /  6.5 & 81.9 / 62.4 & 68.8 /  2.0 & 81.8 / 59.8 & 24.3 /  0.1 & 14.7 /  6.0 & 95.4 / 49.6 & 21.0 / 4.6 & 26.4 / 16.5 \\ 
        \midrule
        ~~VI-Net~\cite{lin2023vi} & \textbf{80.7 / 56.4} & \textbf{90.6 / 79.6} & \textbf{44.8 / 12.7} & \textbf{99.0 / 67.0} & \textbf{96.7 / 72.1} & \textbf{54.9} / \textbf{17.1} & \textbf{52.6} / \textbf{47.3} & \textbf{89.2 / 76.4} &  \textbf{99.1 / 93.7} & \textbf{94.9 / 36.0} & \textbf{85.2 / 62.4}\\
        ~~VI-Net*~\cite{lin2023vi} & 68.4 / 31.3 & 91.1 / 67.7 & 44.1 / 10.2 & 97.7 / 62.9 & 91.8 / 39.0 & 43.0 / 15.6 & 22.8 / 5.8 & 80.7 / 41.4 & 93.7 / 49.2 & 82.2 / 6.3 & 36.5 / 14.5\\
        \bottomrule
    \end{tabular}}
    
    \label{tab:sensorissues}
\end{table*}

\subsection{Scene Statistics}\label{sec:scene_statistics}
\datasetname features 41 large-scale scenes with 194 objects in 10 categories with grasping labels for 16 scenes. It comprises 34 training scenes with 124 objects, 5 test scenes with 50 objects, and 2 validation scenes with 20 objects for object pose estimation tasks. For the 34 training scenes, a total of 20k frames are recorded. Each training scene contains, on average, 6 objects of different categories. The 5 test scenes and 2 validation scenes consist of 3k and 1.4k frames. They are composed of 10 unseen objects per scene with different categories. Compared to other category-level datasets, \datasetname covers the most diverse number of instances and categories. For robotic grasping, we provide 14 training, 1 validation, and 1 test scene. Nonetheless, all 16 scenes can serve to train a real grasping pipeline, as the test would be performed in a real-world setup where success rate serves as the main grasping metric.\par%
\subsection{Viewpoint Coverage}
Established datasets in 6D pose estimation lack well-distributed and dense camera pose coverage around the object. They usually focus on the upper hemisphere, even for large-scale dataset variants like SterOBJ-1M~\cite{liu2021stereobj}. In contrast, \datasetname provides very dense and well-distributed poses (cf. \cref{fig:histograms_comp}). In terms of category level, we compare our trajectories for mutual classes against the recent PhoCal~\cite{wang2022phocal} dataset, which provides very accurate annotations but is limited in the range of motion by the robotic arm used for acquisition (cf. \cref{fig:histograms}).
\section{Benchmark and Experiments}\label{sec:experiments}
\paragraph{Object Pose Estimation} 
In 6D pose estimation, RGB-D input is often used. RGB data aids in classifying objects amidst high intra-class variability. Initially, RGB images identify objects, followed by depth maps for shape and boundary information. NOCS~\cite{wang2019normalized} generates 2D NOCS maps, integrating depth data and ICP for 3D prediction. GPV-Pose~\cite{di2022gpv} segments objects in RGB, then back-projects the depth map for 3D pose prediction through geometry-guided voting. FS-Net~\cite{Chen_2021_CVPR} derives 3D point clouds from depth images post RGB detection, extracting features via a residual network for size and translation estimation. VI-Net~\cite{lin2023vi} simplifies this task, decoupling the rotation into viewpoint and in-plane rotations, learned separately. Our experiment employs Di et al.'s implementation of FS-Net.

\begin{figure}[t!]
 \centering
    \includegraphics[width=1.0\linewidth]{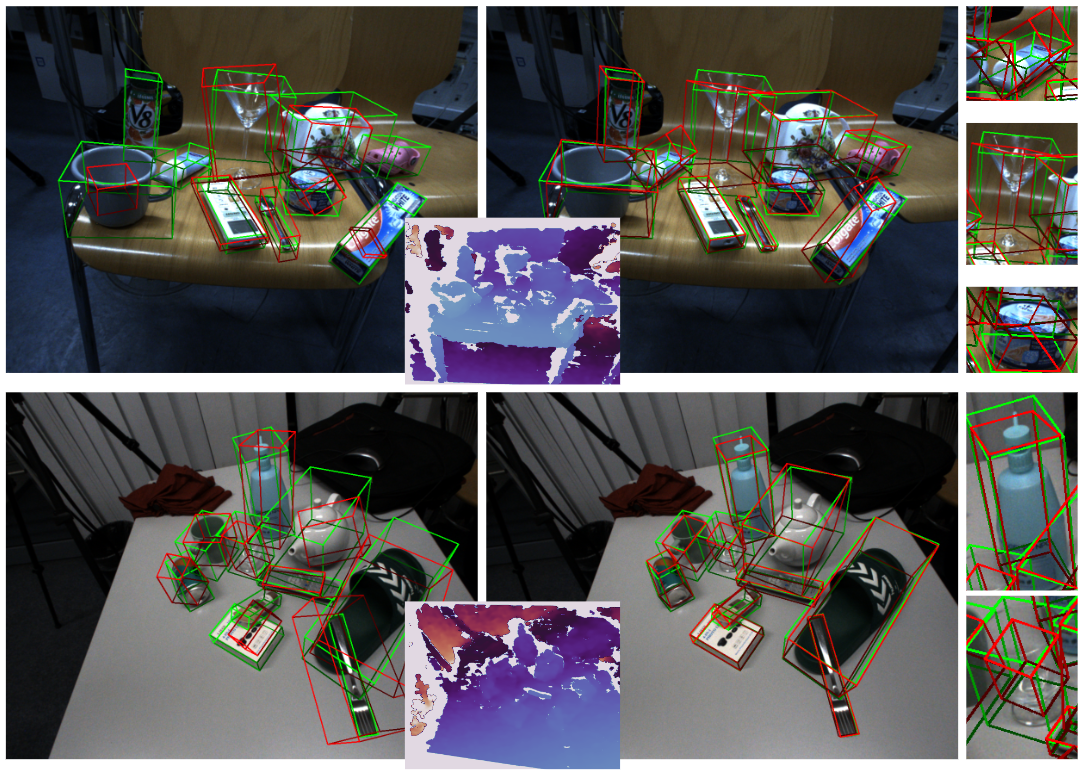}
    \caption{\textbf{Sensor Depth Issues.} Comparison of NOCS prediction (left) and using the NOCS ground truth map but sensor depth (small, center). Even with perfect NOCS maps the lifting from 2D to 3D suffers under the sensor depth map (right).}
    \label{fig:nocs_compare}
\end{figure}

We report IoU results with 25\% and 50\% thresholds in \cref{tab:benchmark} (cf. supp. mat. for additional metrics). The geometry-guided methods GPV-Pose and FS-Net outperform the 2D lifting approach, with VI-Net achieving the best results at both thresholds.
GPV-Pose and FS-Net benefit from precise 2D detection training in HouseCat6D, aiding in detailed object localization. In contrast, NOCS offers a single-stage approach, lifting results from 2D to 3D. 
Compared to the NOCS~\cite{wang2019normalized}, our dataset features cluttered scenes leading to occlusions and closely situated objects (cf. \cref{fig:nocs_compare}). While current methods often overlook occlusions, our initial evaluations suggest significant potential for improvement. The supplementary material details occlusion ratios per category and comparisons with the NOCS dataset. Although VI-Net handles clutter and occlusions better than others, there remains considerable scope for enhancement.

Further, an experiment using NOCS with ground truth predictions but sensor depth alignment reveals inaccuracies in 3D lifting, impacting results as reported in \cref{tab:sensorissues} (top). This is reflected in a significant decline to 22.6\% at the mean IoU at 75\% as also shown in \cref{fig:nocs_compare}. The categories glass and tube suffer especially from the sensor depth. For these, the trained VI-Net even outperforms the ground truth to sensor depth lifting approach.
To demonstrate the importance of extensive pose coverage, we trained VI-Net$^*$ on a data subset similar to the pose coverage of PhoCaL (cf. \cref{tab:sensorissues} (bottom). Results indicate that our scene coverage notably enhances prediction accuracy.

\paragraph{Grasp Pose Estimation} 
KGN~\cite{chen2023keypoint} processes an RGB-D image to estimate gripper keypoints and employs PnP to align 3D keypoints in the gripper frame with 2D camera frame keypoints, solving for 6D grasp poses. We have retrained both the complete KGN model and a simplified version without keypoint offset refinement on HouseCat6D data.

Following the metrics from the original study~\cite{chen2023keypoint}, we report grasp coverage rate (GCR) and object success rate (OSR) in Tab.~\ref{tab:sim} (cf.~\cite{mousavian20196,chu2018real,xu2022gknet}). Real-world experiments were conducted using a 7-DoF Franka robot, with grasp success rates detailed in Tab.~\ref{tab:real}, in line with the approach in~\cite{zhai2022monograspnet} of calculating successful trial percentages over 15 runs per object. Notably, Tab.~\ref{tab:real} demonstrates the method's generalization to unseen objects across different categories. A surprising finding is KGN's efficacy in grasping transparent objects (Glass) post-training on HouseCat6D, both in tests and real-world settings. This contrasts with the original method's limited performance on photometrically challenging objects due to its synthetic training.

\begin{table}[!t]
    \centering
    \caption{\textbf{Grasping Results.} KGN~\cite{chen2023keypoint} \textit{without / with} keypoint refinement, respectively.}
    
    \subcaption{Grasp coverage rate and object success rate on the test set. }
    \resizebox{\columnwidth}{!}{
    \begin{tabular}{l c c c c c c }
    \toprule
    Metric & Bottle & Can & Cup  & Glass & Remote & Tube \\
    \midrule
    GCR (\%) & 17.4 / 24.1 & 35.4 / 58.3 & 32.6 / 34.1 & 16.3 / 40.8 & 64.3 / 64.5 & 47.3 / 61.1 \\
    OSR (\%) & 97.8 / 99.8 & 75.2 / 87.6 & 100 / 100 & 100 / 100 & 92.1 / 94.8 & 100 / 100 \\
    \bottomrule
    \label{tab:sim}
    \end{tabular}
    }

    \subcaption{Real-world grasp success rate.}
    \resizebox{\columnwidth}{!}{
    \begin{tabular}{l c c c c c}
    \toprule
    Metric & Box & Cup & Glass & Remote & Unknown \\
    \midrule
    GSR (\%) & 80.0 / 80.0 & 66.7 / 66.7 & 66.7 / 73.3 & 53.3 / 33.3 & 53.3 / 60.0 \\
    \bottomrule
    \label{tab:real}
    \end{tabular}
    }
    \vspace{-5mm}
\end{table}
\section{Conclusion}
\label{sec:conclusion}
\datasetname is introduced as a comprehensive 6D pose dataset, acquired using a multi-modal camera rig and an external tracking system, offering highly accurate pose annotations. This dataset addresses the limitations of existing datasets by featuring realistic, marker-free scenes with well-distributed object poses. It includes photometrically challenging objects lacking texture and those made of translucent materials, alongside precise robotic grasping annotations. \datasetname, with its quality and breadth, advances research in categorical pose estimation, setting a new standard for applications in everyday household environments and other areas.

\newpage
{
    \small
    \bibliographystyle{ieeenat_fullname}
    \bibliography{main}
}

\newpage
\appendix
\section*{Supplementary}

\begin{figure*}[!t]
    \includegraphics[width=\textwidth]{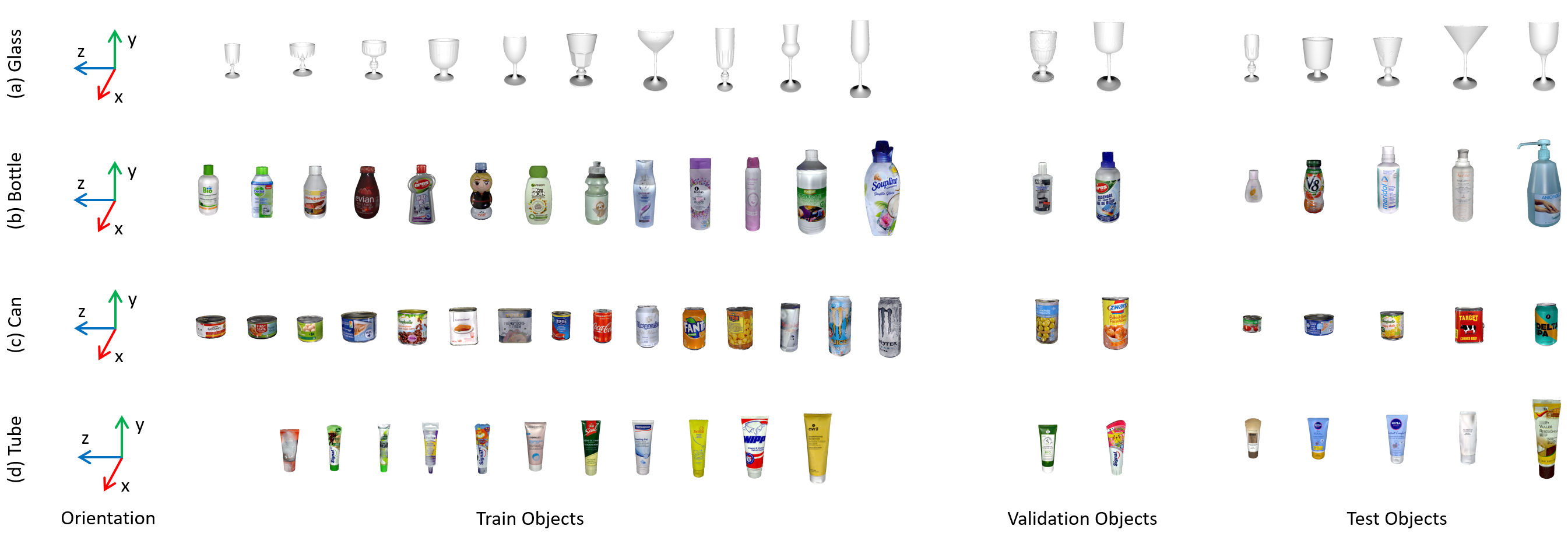}
    \vspace*{-6mm}
    \caption{\textbf{Object Meshes from Symmetric and Partially Symmetric Shape Categories.} Glass (a), bottle (b), can (c), and tube (d) categories are the categories with distinctive symmetry axes. We align the \textit{y} axis to the axis of symmetry. If one surface is larger in area than another side, the \textit{x}-axis is aligned in the perpendicular direction to it. All the objects are rendered in the same scale to highlight the size variance among the same category.}
    \label{fig:partially_symmetric}
\end{figure*}

\section{Object Meshes and Orientation}\label{sec:object_mesh}
The HouseCat6D dataset features 194 highly diverse objects from 10 household object categories with different textures, sizes, and shapes. In this section, we show the meshes of the objects in each category and the descriptions of their orientation.

\paragraph{Glass}
HouseCat6D aligns the symmetry axis with the \textit{y} axis for the (partially) symmetric objects. Glass objects in our dataset are fully symmetric around \textit{y} axis in accordance with~\cite{wang2019normalized} who also align \textit{y} axis and symmetry axis. The \textit{x} and \textit{z} axes serve as any orthogonal axes around the \textit{y} axis as exemplified in \cref{fig:partially_symmetric} (a).

\paragraph{Bottle}
Unlike the glass objects, bottle objects in our dataset are sometimes not fully symmetric (i.e. frontal surface is wider than the side) as in \cref{fig:partially_symmetric} (b). In this case, we define the \textit{x} axis perpendicular to the surface of larger area. 

\paragraph{Can}
Similar to the bottle objects, can objects in our dataset sometimes are not fully symmetric (i.e. some cans are more square and one side is wider than the other side) as shown in \cref{fig:partially_symmetric} (c). Like the bottle objects, we define the \textit{x} axis perpendicular to the wider side. 

\paragraph{Tube}
Tube objects in our dataset are partially symmetric in shape, such that one side is round at the end while flat on the other side as shown in \cref{fig:partially_symmetric} (d). As in the can and bottle category, we define the \textit{x} axis perpendicular to the wider side. 

\begin{figure*}[t]
 \centering
    \includegraphics[width=1.0\linewidth]{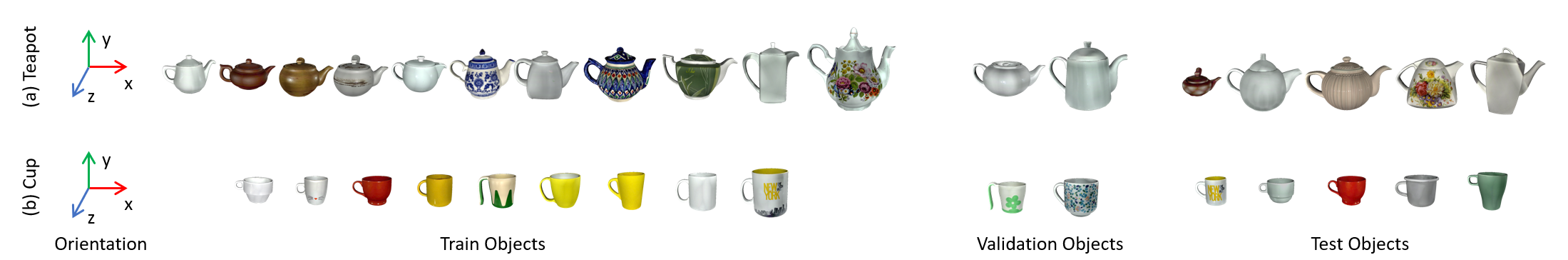}
    \caption{\textbf{Object Meshes from (Partially) Symmetric Objects With a Handle.} Teapot (a) and cup (b) are the categories with objects that include a (partially) symmetric body with handle. We align the \textit{y} axis with the symmetry axis of the body and the \textit{x} axis with the direction from handle to the the other side of the body. All the objects are rendered in the same scale to highlight the size variance among the same category.}
    
    \label{fig:handle_objects}
\end{figure*}

\paragraph{Teapot}
In general, teapots have the shape of one (partially) symmetric body with a handle and tip where the liquid comes out. In our dataset, we use the \textit{y} axis for the direction of the symmetric body and \textit{x} axis for the direction from the handle to the tip as shown in \cref{fig:handle_objects} (a).

\begin{figure*}[t]
 \centering
    \includegraphics[width=1.0\linewidth]{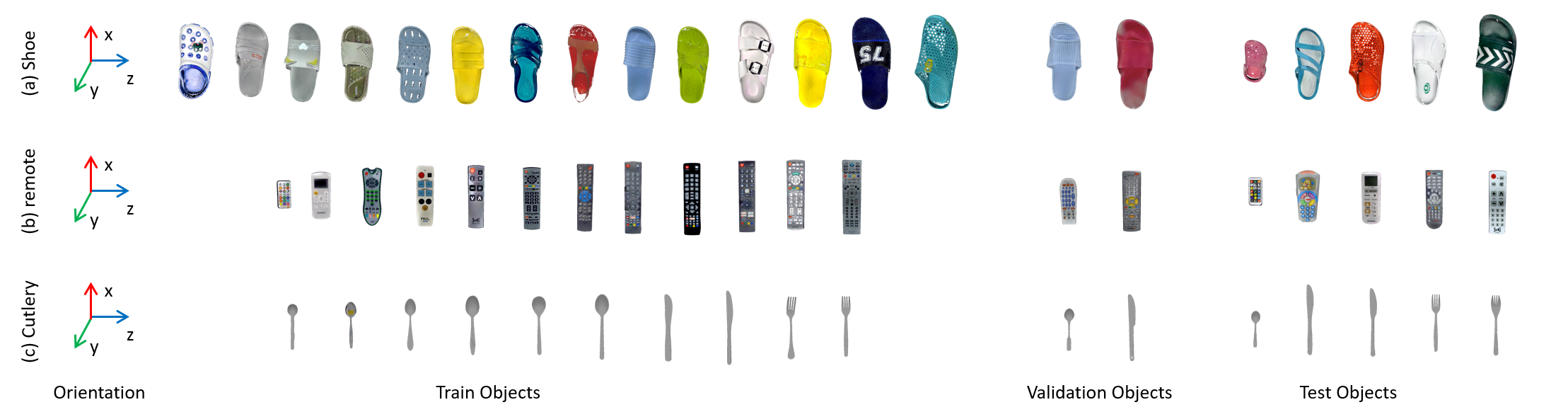}
    \caption{\textbf{Object Meshes from Flat Shape Categories}. Shoe (a), remote (b) and cutlery (c) are the categories with long, flat and non-symmetric shape. We oriented such shapes in a way that the \textit{y} axis points in the direction of the upper side and \textit{x} in the direction of the front side. All the objects are rendered in the same scale to highlight the size variance among the same category.}
    
    \label{fig:flat_objects}
\end{figure*}
\begin{figure*}[!ht]
 \centering
    \includegraphics[width=1.0\linewidth]{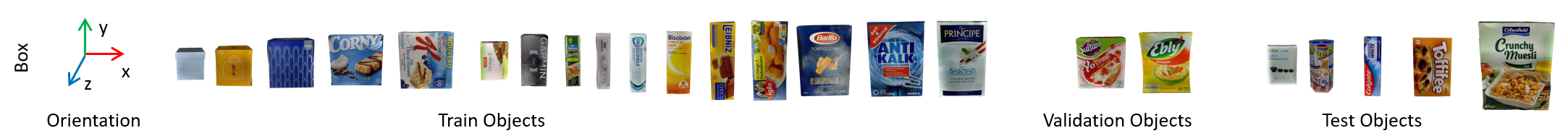}
    \caption{\textbf{Object Meshes for Box category.} Unlike the other categories, the sides of the box are rather defined by their texture. To allow networks to generalize in this category, we orient the meshes by their side length. We set \textit{y}, \textit{x}, \textit{z} as direction of first, second and third longest side. All the objects are rendered in the same scale to highlight the size variance among the same category.}
    
    \label{fig:box_objects}
\end{figure*}

\paragraph{Cup}
For the cup category, we only use cups with handles that have the shape of one symmetric body with a handle. Thus, similar to the Teapot category, we align the \textit{y} axis to the direction of the symmetric body and \textit{x} with the direction from the handle to the other side of the body as shown in \cref{fig:handle_objects} (b).

\paragraph{Shoe}
Shoes, in general, have a long, flat and non-symmetric shape. For this category, we use only the right side of the slipper as illustrated in \cref{fig:flat_objects} (a). We oriented shoes such that their upper side points in the direction of the \textit{y} axis and the front side points in the direction of the \textit{x} axis.

\paragraph{Remote}
Remotes have relatively flat bodies with long and non-symmetric shapes, as shown in \cref{fig:flat_objects} (b). Similar to the shoe category, remotes are oriented such that their upper side points in the direction of the \textit{y} axis, and the front side is is oriented in the direction of the \textit{x} axis.

\paragraph{Cutlery}
Although the texture of the reflective surface makes a clear distinction between the cutlery category to any other category, the shape itself shares similarity with shoe and remote category. It is flat, long, and non-symmetric (\cref{fig:flat_objects} (c)). Thus, it shares the same orientation scheme, the upper side is aligned with the \textit{y} axis and the front side points in \textit{x} direction.

\begin{figure*}[!t]
 \centering
    \includegraphics[width=1.0\linewidth]{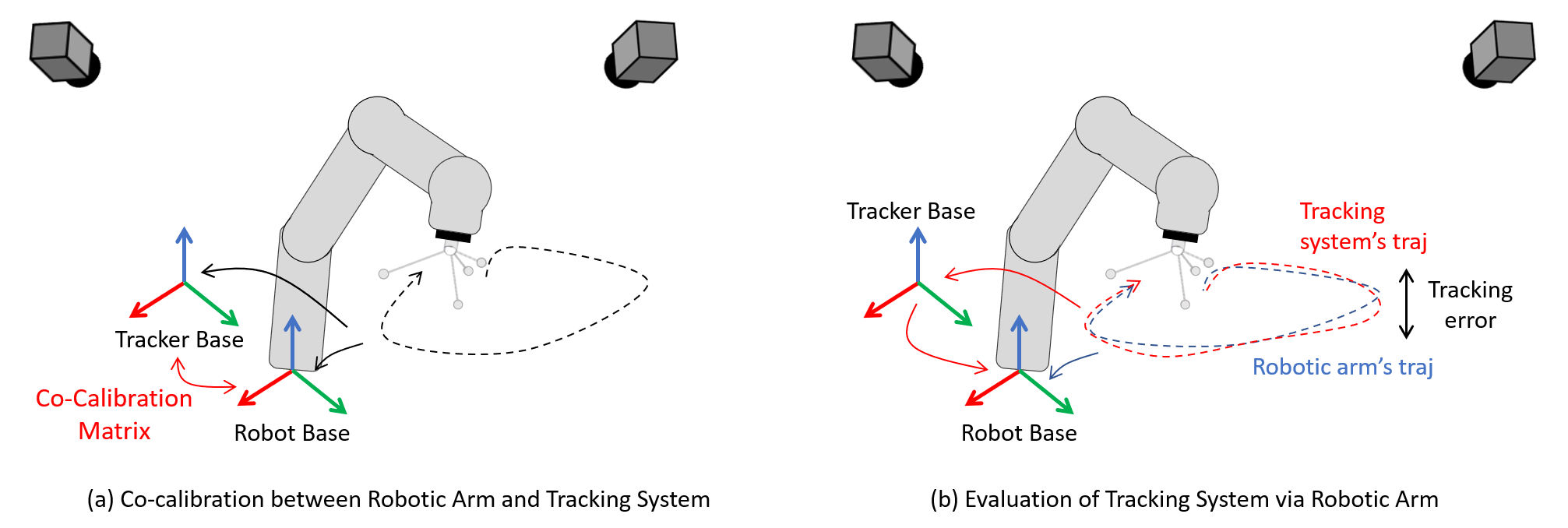}
    \caption{\textbf{Tracking System Evaluation.} We use a robotic arm to evaluate the quality of the tracking system. We first (a) co-calibrate the robot and the tracking system such that they share a common reference frame and then (b) run an example trajectory to calculate the difference between the trajectory obtained from the robot and the tracking system for error evaluation.}
    \label{fig:tracking_evaluation}
\end{figure*}

\paragraph{Box}
Unlike other categories, the sides of the box are defined by their texture. Even a human observer has to inspect the textures on multiple sides of a box to judge which side is the front or upper side \etc. To make it easier for networks to generalize the orientation of boxes, we orient them by the length of the sides independent of their textures. We use \textit{y}, \textit{x}, \textit{z} for the direction of the first, second, and third longest side as shown in \cref{fig:box_objects}.

\section{Hardware Details} \label{sec:hardware_detail}
In this section, we provide detailed information about the hardware we used for the dataset acquisition.

\paragraph{3D Scanning}
As shown in Sec.\ref{sec:object_mesh}, our dataset comprises of 10 household categories such as bottle, box, can, cup, cutlery, glass, remote, shoe, teapot, tube. To ensure the high quality meshes we use 3D scanner equipped with turn table and structured light stereo system (EinScan-SP 3D Scanner, SHINING 3D Tech. Co., Ltd., Hangzhou, China), which produces single shot accuracy of $\leq 0.05~\text{mm}$ in a scanning volume of $1200 \times 1200 \times 1200~\text{mm}^3$. For photometrically challenging categories like cutlery and glass, self-vanishing 3D scanning spray (AESUB Blue, Aesub, Recklinghausen, Germany) is applied prior to the scanning.
\paragraph{External Tracking System.}
To ensure broad viewpoint coverage with high-quality annotation without using a checkerboard, we utilize an external tracker system composed of 4 (2x Stereo) ARTTRACK2 cameras (Advanced Realtime Tracking GmbH \& Co, Germany) with built-in infrared flash (NIR, 880 nm) and maximum tracking distance of 4.5~m for both object pose and camera pose annotation. 
\paragraph{Cameras.} Our multi-modal dataset comprises two main modalities: Polarimetric RGB image and active stereo depth. A Phoenix 5.0 MP Polarization camera with Sony IMX264MYR CMOS Polarsens (PHX050S1-QC, LUCID Vision Labs, Inc., Canada) sensor is used to produce the RGB+P images, and Intel RealSense D435 (RealSense D435i, Intel, USA) acquires the depth maps. We specifically choose D435 as the depth sensor over Time-of-Flight sensors as active stereo depth provides, in general, more robust depth on photometrically challenging material~\cite{hammer}. To ensure the best synchronization between the two cameras, we use an external tracking signal provided by a Raspberry Pi (Raspberry Pi Foundation, United Kingdom) with GPIO output and later use the trigger signal as the timestamp of images for post-ex synchronization correction with the tracking system.

\section{External Tracking System Evaluation} \label{sec:external_tracking_eval}
As mentioned in Sec.~3.2 in the main paper, we evaluate our IR-based external tracking system ARTTRACK2 via a robotic arm. We use a KUKA LBR iiwa 7 R800 (KUKA Roboter GmbH, Augsburg, Germany), a 7 DoF robotic arm certified for industrial use to provide $\pm0.1$~mm positional reproducibility, as the device to produce the ground truth pose for the comparison. In this section, we describe the detailed steps for the evaluation.

\subsection{Robot-Tracker Co-Calibration} \label{subsec:cocalibration}
The first step to evaluate the tracking system with a robot is to co-calibrate the base of the robot and the tracking system. For this, we attach the calibrated IR tracking body on the robotic End-Effector (EE) as shown in Fig.~\ref{fig:tracking_evaluation} (a). We then acquire one trajectory from two different coordinate bases, one from the Robot base and the other one from the Tracker base. Similar to hand-eye calibration, we extract the static transformation between the two trajectories using the method of Horn~\cite{horn}. In this case, the static transformation matrix is the transformation between Tracker Base and Robot Base (marked red in Fig.~\ref{fig:tracking_evaluation} (a)).

\subsection{Trajectory Error Evaluation} \label{subsec:trajectory_error}
After co-calibration, we keep the tracking body on the robotic EE and make an evaluation trajectory that replicates the trajectory in one of the scenes. We repeat the trajectory twice, once with the robot stopping at every capturing position and once with the robot not stopping during the pose capture. The first trajectory serves as an evaluation for the tracking system accuracy in the static case, and the later trajectory serves as an evaluation in the dynamic case. As it is possible to obtain the pose of the tracking body from both, robot and tracking system, in the same coordinate frame using the co-calibration matrix, the error of the tracking system is calculated as the pose difference between the pose from the robotic arm and the pose from the tracking system (Fig.~\ref{fig:tracking_evaluation} (b)). We measure an error of $0.67$~mm / $0.12^{\circ}$ in the static case and $0.92$~mm / $0.16^{\circ}$ in the dynamic case.
\begin{figure}[t]
 \centering
    \includegraphics[width=0.5\linewidth]{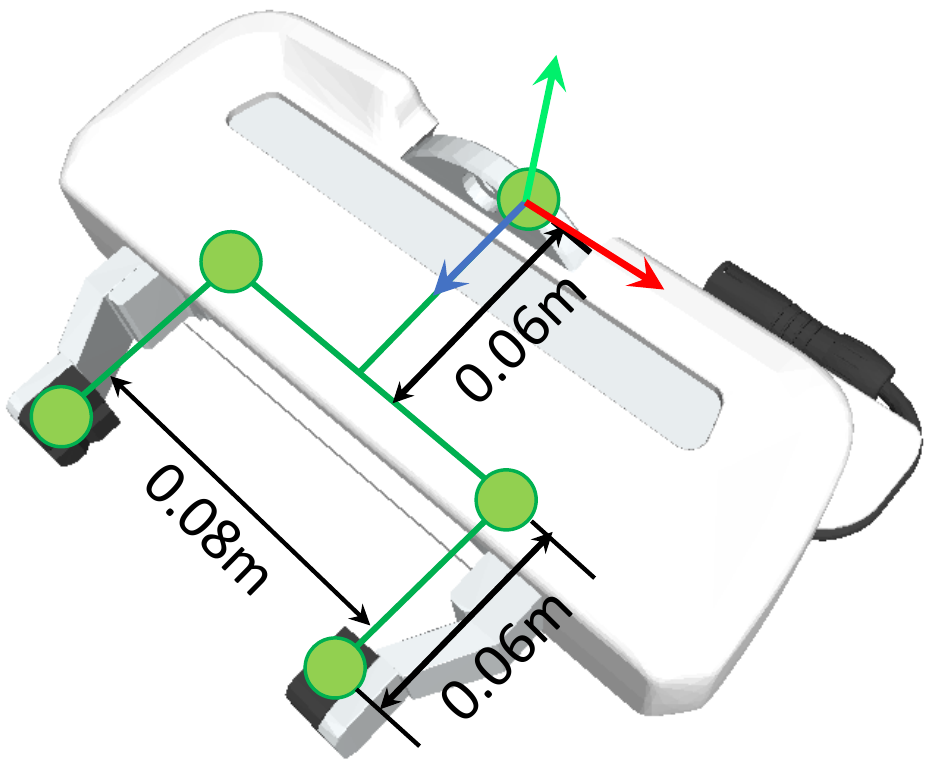}
    \caption{\textbf{The model of the parallel-jaw gripper}, whose finger depth is $0.04m$, maximum grasping width is $0.08m$, and the distance between the gripper base and the center of two fingers' base is $0.04m$. }
    \label{fig:gripper}
\end{figure}
\begin{figure}[t]
 \centering
    \includegraphics[width=1.0\linewidth]{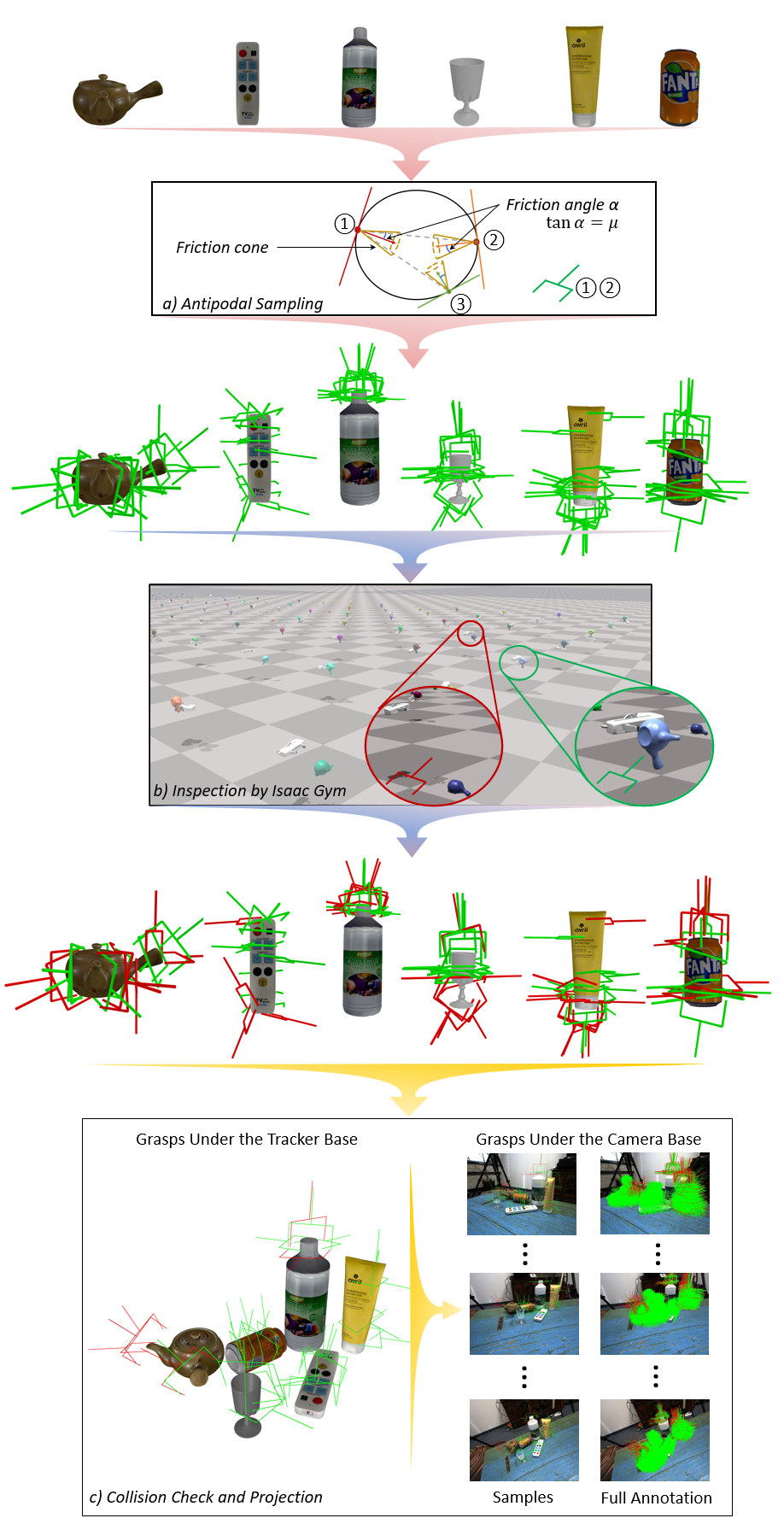}
    \caption{\textbf{The pipeline of the grasp annotation process.} We show downsampled grasps for better visualization and show the full annotation at the end for the final performance.}
    \label{fig:grasp_anno}
\end{figure}
\section{Grasping Annotation Pipeline} \label{sec:grasping_annotation}
In this section, we detail the grasping annotation process. The pipeline is illustrated in \cref{fig:grasp_anno}. %
For each scene, we first obtain the scene by reconstructing the background (e.g. table) with multiview depth and displacing the object meshes on the top of the background mesh according to their pose. After successfully reconstructing the scene, the meshes are sent to the antipodal sampling module to generate grasp candidates (\cref{fig:grasp_anno}.a).
Then Isaac Gym~\cite{makoviychuk2isaac} sorts out the good grasps among all candidates for each object by checking if grasping an object failed. Successful grasps are in green, while failed grasps are in red (\cref{fig:grasp_anno}.b). Then objects are projected to the tracker base along with their associated grasps to check the collisions and collided grasps are removed from the original ones. Finally, we project these checked grasps to each image base to obtain the ultimate dataset. (\cref{fig:grasp_anno}.c). 

\subsection{Scene Mesh Acquisition} 
\begin{figure}[t]
 \centering
    \includegraphics[width=1.0\linewidth]{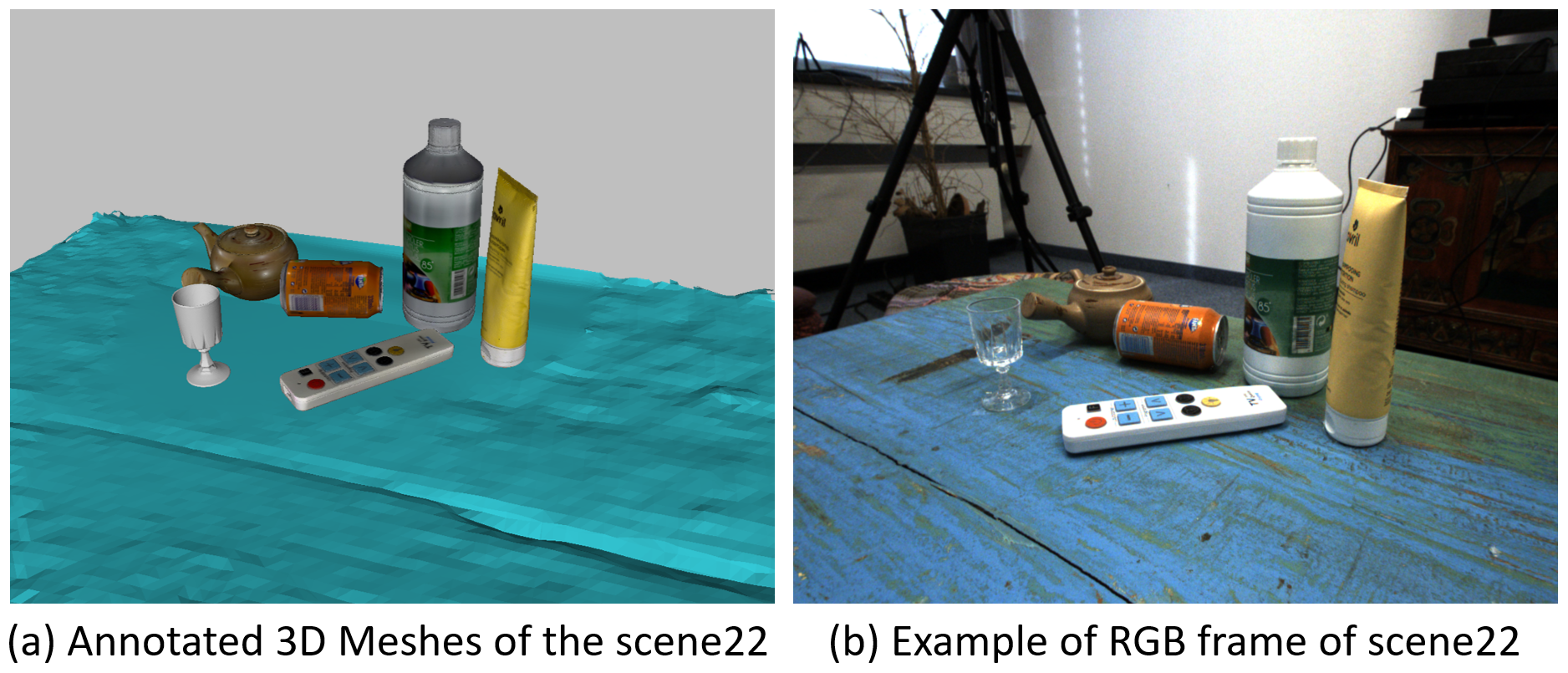}
    \caption{\textbf{Example of Mesh Annotation and Its Corresponding RGB Image.} We annotate the scene by reconstructing the platform and displacing the objects' meshes with their pose. Note that we do not reconstruct the other parts of the background such as the wall as they are not necessary for grasping simulation.}
    \label{fig:3d_annotation}
\end{figure}

To annotate the correct grasping position with collision inspection, it is important to have a full mesh of the scene, which contains objects as well as their platform where the object are placed, such that physical simulation can filter out the grasping points which leads collision of gripper on the other objects and the background. For the objects, we displaced their meshes in the scene with the annotated poses. On the other hand, for the platform, it is not possible to do the same way as the background is not scanned prior. Instead, we reconstruct the scene with the depth images with the corresponding camera poses using truncated signed distance fusion and hole followed by manual hole filling with Artec Studio 17 Professional (Artec3D, Senningerberg, Luxembourg). An example of the 3D mesh of objects with the reconstruction of the platform is shown in \cref{fig:3d_annotation} with an example of an RGB frame from the corresponding scene.

\subsection{Antipodal Sampling}
Antipodal sampling is a wide-used technique for grasp pose generation, which has been investigated in several previous works~\cite{mahler2017dex,eppner2021acronym,zhai20222}. Given an object mesh, this scheme first samples an arbitrary point on the mesh surface as the initial contact point (\ding{172}) together with a line within a range around the surface normal. The sampling threshold $\mu = \tan(\alpha)$ with the friction angle $\alpha$ restricts the range at which rays can be emitted. A second point (\ding{173} / \ding{174}) is found as the intersection of both mesh and line. Then reject sampling is used to prune the point whose line is not inside the friction cone (\ding{174}) or whose distance from the initial point is beyond the max width of the gripper model. A successfully sampled grasp ${{ G}}_{obj}$ is then derived by taking the center point between two contact points (\ding{172}\ding{173}) and a randomly sampled rotation around the line. Here, in this work, we set $\mu$ as $0.4$. The end-effector model we use is a Franka Emika parallel-jaw gripper, as shown in \cref{fig:gripper}.

\subsection{Simulation Inspectation}
After obtaining grasp samples, we use a physical engine, namely Isaac Gym, to inspect grasps which are successful. For each object, we parallelly create the same number of simulation environments as of grasps belonging to the object. We inspect whether these grasps are successful by calculating the distance between the gripper and the centroid of the object model 15 seconds after the finger closure defined by individual grasping width. If the distance is less than $0.1m$, we label this grasp as a successful one and vice versa.
\subsection{Grasp Projection}
This is a two-stage procedure. We retrieve objects in each scene and replicate the first-stage projection for all objects in the scene, where we transform the grasps belonging to an object to the tracker base according to the object pose and check their collisions with the surrounding meshes, including other objects and the background. The collision checking module is from the public library Trimesh\footnote{\url{https://trimsh.org/trimesh.collision.html}}. Then we project all grasps to every image frame to obtain the final dataset, utilizing the camera trajectory recorded under the tracker base.
\section{Occlusion Analysis} \label{sec:occlusion_analysis}
\begin{figure}[t]
 \centering
    \includegraphics[width=1.0\linewidth]{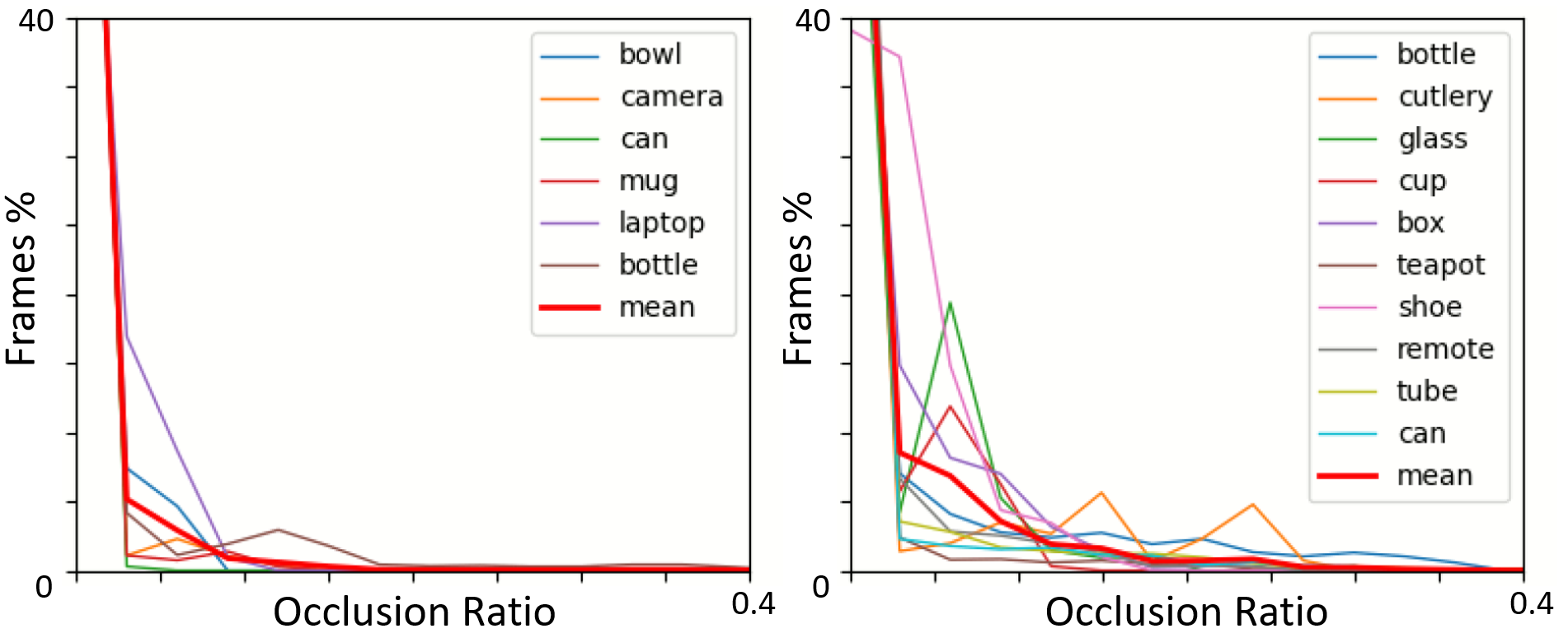}
    \caption{\textbf{Occlusion Comparison between NOCS~\cite{wang2019normalized} and HouseCat6D.} HouseCat6D covers more occlusions as well as more frequency on the occlusion, which makes the dataset more challenging as well as closer to the real-life scenario.}
    \label{fig:occlusion_plot}
\end{figure}

When it comes to detecting the objects and estimating the pose, occlusion and visibility take important roles. In our dataset, we provide the visibility ratio of each category in the scene per frame. The visibility is calculated as follows. Firstly, we render the mask of an object with a given pose, one object per time to prevent occlusion between categories, and count number of pixels in the mask \(M^{cat}_{full}\).
Then masks of categories are rendered again but all together so that occlusion is accounted, followed by counting the number of pixels on each object \(M^{cat}_{occluded}\), which now has fewer pixels due to occlusion from other objects. Occlusion ratio is calculated as \(M^{cat}_{occluded}\) / \(M^{cat}_{full}\), which then averaged over all frames and scenes. We show the ratio on our dataset and as well as on NOCS dataset~\cite{wang2019normalized} in Fig.~\ref{fig:occlusion_plot} to emphasize the difference in terms of the occlusion in the dataset.

\begin{figure}[t]
 \centering
    \includegraphics[width=1.0\linewidth]{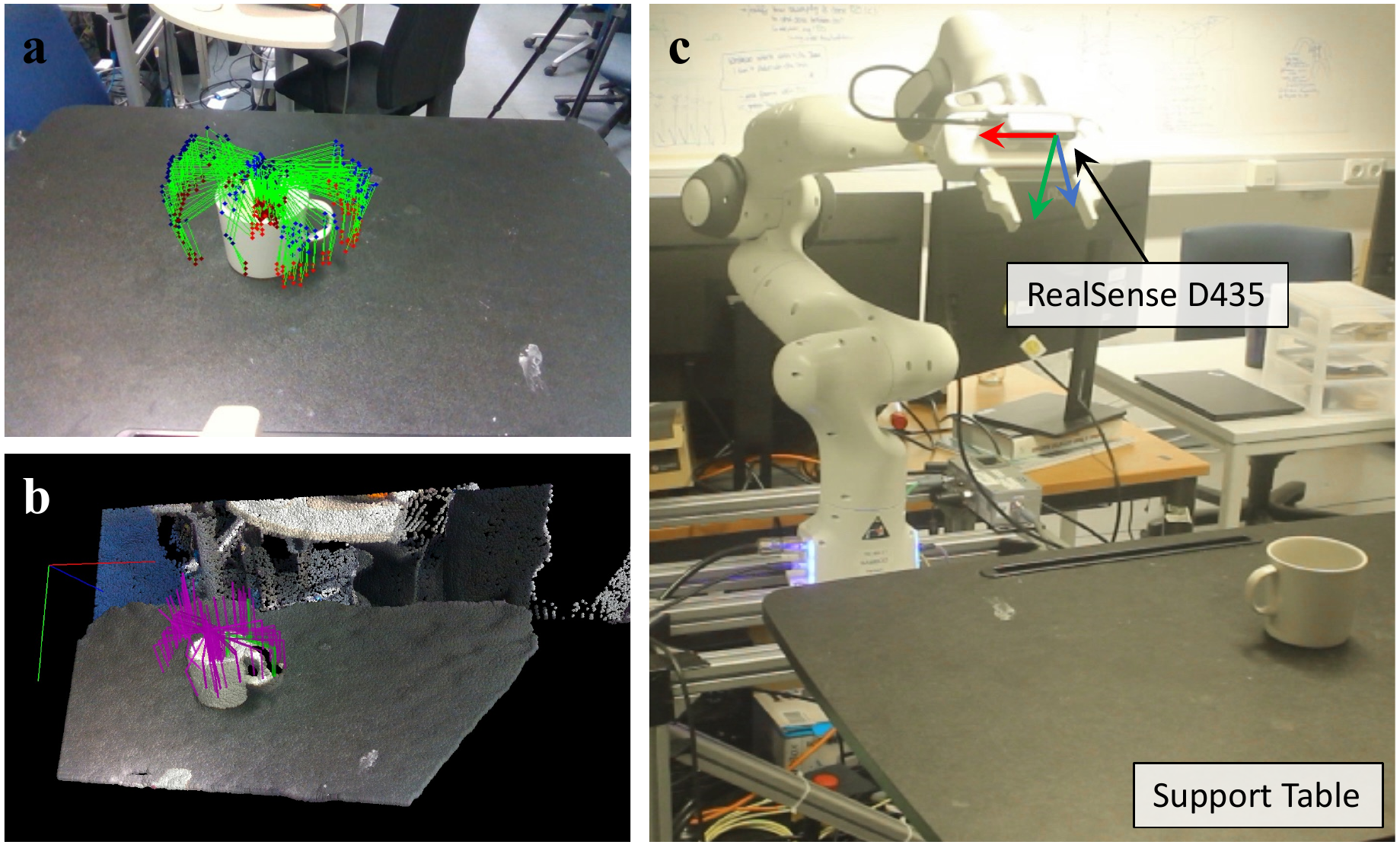}
    \caption{\textbf{An example of a real-world grasping trial.} (a) 2D grasp keypoints in the camera view. (b) 3D grasps visualization, with the best in green and the rest in purple. (c) Hardware setup in the third-person view.}
    \label{fig:real_grasp}
\end{figure}
\section{Evaluation on Rotation Translation Metric} \label{sec:rotation_translation}

\begin{table}[!ht]
\caption{\textbf{Quantitative Evaluation on Rotation and Translation Metric.} For rotation and translation metric, we show average accuracy over all categories.}
    \centering 
    \resizebox{0.85\columnwidth}{!}
    {
    \begin{tabular}{l|cccc}
     Threshold &  NOCS~\cite{wang2019normalized} & GPV-Pose~\cite{di2022gpv} & FS-Net~\cite{Chen_2021_CVPR} & VI-Net~\cite{lin2023vi}\\ \midrule
     $10^\circ5$cm & 4.8 & 22.7 & 21.6 & 29.1 \\
     \bottomrule
    \end{tabular}}
    
    \label{tab:benchmark_rot_trans}
\end{table}

In \cref{tab:benchmark_rot_trans}, we show the evaluation of baseline on rotation and translation error metric with a set of thresholds: 10\textdegree5cm. Similar to 3D IOU, NOCS~\cite{centersnap} performs significantly worse in our dataset compared to other baselines. As mentioned in the main paper, we suspect this is due to issue with inaccurate depth being used for lifting NOCS~\cite{centersnap} prediction in 2D into 3D. On the other hand, geometric guided approach such as GPV-Pose~\cite{di2022gpv}, FS-Net~\cite{chen2021fs} and VI-Net~\cite{lin2023vi} has better performance with ground truth detection mask. Especially, with better parameterization on rotation, VI-Net~\cite{lin2023vi} performs significantly better compared to other geometric approach, GPV-Pose~\cite{di2022gpv} FS-Net~\cite{chen2021fs}.
\section{Real-World Grasping} \label{sec:grasping}
Unlike the experiments in simulation which are conducted on the available test set, real-world grasping is more challenging with respect to two facts. First, the objects are more random and are not included in the dataset, with some of them even in the unseen categories, which tests the generalization ability of the network. Second, the appearance of the backgrounds are more complex and the imaging style is also different since the imagery sensor is different from the one collecting the dataset, which tests the robustness of the network.

\paragraph{Hardware Setup}
We test the trained KGN~\cite{chen2023keypoint} in real-world scenarios using a 7-DoF Franka Panda robot with a parallel-jaw gripper as the end-effector. The sensor mounted on the gripper base is a RealSense D435 RGB-D camera. The framework is run on an NVIDIA A4000 GPU.
\paragraph{Implementation Details}
We randomly select support tables with unseen backgrounds as the grasping environment. Then we fix a certain sequence of joint positions for the robot as the home position where the camera observes the table from the side, as shown in Fig.~\ref{fig:real_grasp}.c. We select three types of objects for the test--1) normal objects in the seen categories, 2) normal objects in the unseen categories, and 3) photometrically challenging objects in the seen categories, whose grasp success rates are reported in the main paper. For example, in the first type, we let the robot grasp a cup, shown in Fig.~\ref{fig:real_grasp}. KGN~\cite{chen2023keypoint} starts to infer 2D grasp keypoints on the image (Fig.~\ref{fig:real_grasp}.a), then it utilizes PnP and 3D keypoints shown in Fig.~\ref{fig:gripper} with camera intrinsics to solve 3D grasp poses (Fig.~\ref{fig:real_grasp}.b).
\section{Ablation Study} \label{sec:ablation_styudy}

\begin{table*}[!t]
\caption{\textbf{Ablation Study on Different Input} Class-wise evaluation of 3D IoU (at 25\%/ at 50\%)  for VI-Net~\cite{lin2023vi} with different training setup.
    }
    \centering
    \resizebox{\textwidth}{!}{ 
    \begin{tabular}{l|c|ccccccccccc} \toprule
       Approach & Train Set & $\text{3D}_{25}$ / $\text{3D}_{50}$ & Bottle & Box & Can & Cup & Remote & Teapot & Cutlery & Glass & Tube & Shoe \\
       \midrule
        \multirow{3}{*}{VI-Net~\cite{lin2023vi}} 
        & Full & 80.7 / 56.4 & 90.6 / 79.6 & 44.8 / 12.7 & 99.0 / 67.0 & 96.7 / 72.1 & 54.9 / 17.1 & 52.6 / 47.3 & 89.2 / 76.4 &  \textbf{99.1 / 93.7} & \textbf{94.9 / 36.0} & 85.2 / 62.4\\
        & RV & 74.2 / 46.8 & 91.0 / 76.6 & 59.1 / 23.5 & 98.9 / 67.2  & 76.0 / 36.6  & 59.4 / 34.3  & 22.7 / 18.8  & 79.4 / 57.3  & 97.7 / 85.3  & 66.3 / 47.8 & 91.4 / 20.4 \\
        & RS & 67.7 / 35.8  & 90.1 / 68.7 & 49.0 / 9.8 & 96.9 / 53.6 & 87.2 / 48.5 & 40.2 / 16.3  & 28.8 / 15.8 & 67.4 / 49.0 & 98.5 / 73.6 & 86.6 / 7.9 & 32.4 / 14.9\\
         
        \bottomrule
    \end{tabular}}
    
    \label{tab:benchmark_ablation}
\end{table*}

We trained VI-Net~\cite{lin2023vi} on our dataset with different setups, such as reduced viewpoint coverage of camera (RV), reduced number of scenes (fewer objects per category) (RS) to study the impact of different aspect of the dataset on category level 6d pose estimation task. The results are summarized in Tab.~\ref{tab:benchmark_ablation}. For RV and RS setup, we specifically mimic the coverage of PhoCal~\cite{wang2022phocal} by using less number of scenes (RS) and selecting the subset of camera trajectory as continuous 250 frames of translation-dominated motion (RV).

\paragraph{Impact of View VS Scenes} Compared to having reduced viewpoints (RV) during training, reducing the scene (RS) has a more negative impact on the test evaluation. As the main task of category-level pose estimation is about generalizing on the unseen objects of known categories, we find it beneficial to see more objects and backgrounds even if the viewpoint is limited. This further highlights the advantage of our dataset over NOCS dataset~\cite{wang2019normalized} and PhoCal dataset~\cite{wang2022phocal} for both the number of scenes and the number of objects. Furthermore, when both RS and RV are combined, there is a significant drop in the performance, which gives an advantage of our dataset over PhoCal~\cite{wang2022phocal}, where the robotic arm annotations have a clear limitation on the viewpoint coverage as well as the number of scenes.

\section{Dataset Sample} \label{sec:Dataset}
\begin{figure*}[!ht]
 \centering
    \includegraphics[width=1.0\linewidth]{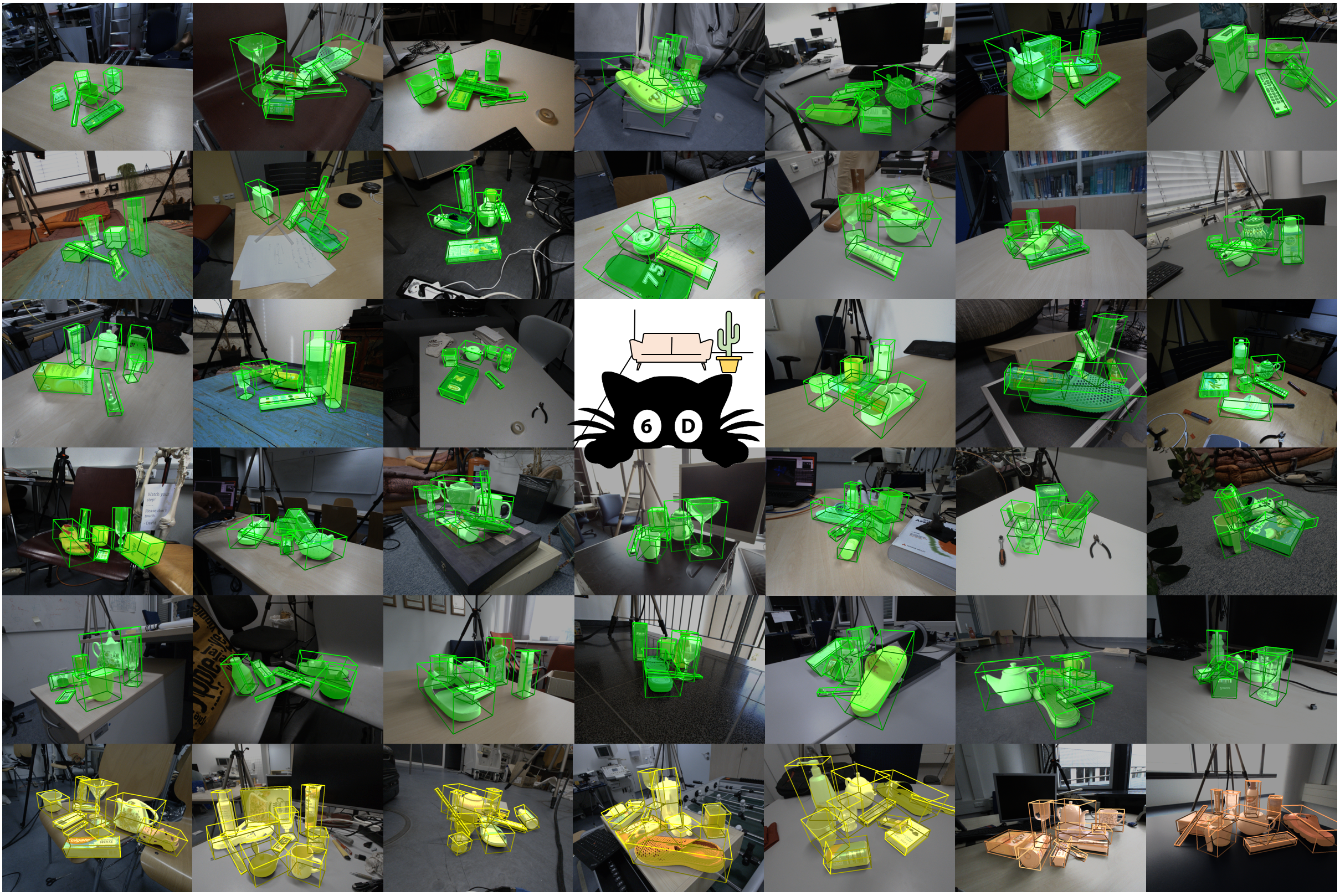}
    \vspace{-4mm}
    \caption{\textbf{Dataset Sample.} Our dataset is composed of 41 scenes with high quality annotations structured in 34 training scenes (green), 5 test scenes (yellow), and 2 validation scenes (orange). We overlay rendered object masks as well as bounding boxes to highlight the quality of our dataset annotation.}
    
    \label{fig:dataset_example}
\end{figure*}

Fig.~\ref{fig:dataset_example} shows example images of our dataset from all 41 scenes. In Fig.~\ref{fig:dataset_example}, we augment rendered object masks together with bounding boxes to highlight the quality of our dataset annotation. Training scenes are augmented with green, test scenes are augmented with yellow, validation scenes are augmented with orange color.

\end{document}